\documentclass[10pt,twocolumn,letterpaper]{article}

\usepackage{cvpr}
\usepackage{times}
\usepackage{epsfig}
\usepackage{graphicx}
\usepackage{amsmath}
\usepackage{amssymb}
\usepackage{stmaryrd}
\usepackage{algorithm}
\usepackage{algorithmic}
\usepackage{subfigure}
\usepackage{tabularx}
\usepackage{tabularx}
\usepackage{multirow}
\usepackage{array}
\usepackage{enumerate}
\usepackage{paralist}
\usepackage{tabularx}

\newcolumntype{L}[1]{>{\raggedright\let\newline\\\arraybackslash\hspace{0pt}}m{#1}}
\newcolumntype{C}[1]{>{\centering\let\newline\\\arraybackslash\hspace{0pt}}m{#1}}
\newcolumntype{R}[1]{>{\raggedleft\let\newline\\\arraybackslash\hspace{0pt}}m{#1}}

\newcommand{\argmin}{\operatornamewithlimits{argmin}}

\newcommand*\rot{\rotatebox{90}}

\newcommand*\concat{\mathbin{\|}}

\usepackage[pagebackref=true,breaklinks=true,colorlinks,bookmarks=false]{hyperref}

\makeatletter
\def\hlinewd#1{%
  \noalign{\ifnum0=`}\fi\hrule \@height #1 \futurelet
   \reserved@a\@xhline}
\makeatother

\cvprfinalcopy 

\def\cvprPaperID{93} 

\begin{document}

\title{Self-Supervised Learning for Stereo Matching with Self-Improving Ability}

\author{Yiran Zhong $^{1,3}$,   Yuchao Dai $^{1}$,   and   Hongdong Li  $^{1,2}$\\
$^{1}$Australian National University, $^{2}$ Australian Centre for Robotic Vision, $^{3}$ Data61\\ 
{\tt\small\{yiran.zhong, yuchao.dai, hongdong.li\}@anu.edu.au}
}

\maketitle

\begin{abstract}

Exiting deep-learning based dense stereo matching methods often rely on ground-truth disparity maps as the training signals, which are however not always available in many situations.  In this paper, we design a simple convolutional neural network architecture that is able to learn to compute dense disparity maps directly from the stereo inputs.  Training is performed in an end-to-end fashion without the need of ground-truth disparity maps.  The idea is to use image warping error (instead of disparity-map residuals) as the loss function to drive the learning process, aiming to find a depth-map that minimizes the warping error. While this is a simple concept well-known in stereo matching, to make it work in a deep-learning framework, many non-trivial challenges must be overcome, and in this work we provide effective solutions.  Our network is self-adaptive to different unseen imageries as well as to different camera settings. Experiments on KITTI and Middlebury stereo benchmark datasets show that our method outperforms many state-of-the-art stereo matching methods with a margin, and at the same time significantly faster.

\end{abstract}

\section{Introduction}


This paper is concerned with the classic problem of stereo matching, {\em i.e.} computing a dense depth/disparity map from a pair of stereo images.  This problem has been extensively studied, yet recent advent of deep learning has provided new solutions with unprecedented state-of-the-art performance both in accuracy and in efficiency.  Currently, the leading stereo methods in almost all popular benchmarks (\eg, KITTI dataset \cite{Geiger2012CVPR}, Middlebury dataset \cite{Scharstein2002}) are deep-learning based.  However, most of these deep stereo matching methods crucially rely on the availability of proper ground-truth depth-map labellings to be used as the training signals in network learning. As is well known, capturing ground truth depth maps is a laborious task, not always possible, and often plagued with noise as well.

In contrast, traditional stereo matching methods (\eg, max-flow \cite{Kolmogorov2001}, belief propagation \cite{Klaus2006}, semi-global matching \cite{Hirschmuller2008}) do not need ground-truth depth-maps (other than in meta-parameter tuning stage during cross validation). Traditional stereo matching methods generally consist of four steps: matching cost computation, cost aggregation, optimization, and disparity refinement, where each module is carefully designed manually. In principle, all these modules can be realized by using deep neural network, without the explicit need of ground-truth depth-maps. 

In this work, we demonstrate that one can train an end-to-end deep stereo matching network without ground-truth depth maps as the training signals and thus derive an \emph{self-supervised learning} framework to stereo matching. We show the stereo image warping errors themselves (left to right, and right to left) are sufficient to drive a deep network to converge to the right state that leads to superior stereo matching performance, even on never-seen-before stereo imageries. 

Whist the basic idea may seem trivial, to achieve this one has to overcome several design difficulties or barriers in both network design and loss function selection. Specifically, because the network training is only based on photometric errors between the left and right images, there could be multiple possible solutions that minimize the warping error. To overcome this, we propose to use 3D regularization in the high-dimension feature volume to push away those trivial solutions. We choose the disparity map which achieves the minimal distance in the convolutional feature space as well as in the appearance space. In addition, a novel left-right consistency check loss function is proposed to effectively handle the textureless regions. We will explain these in details later.


\begin{figure*}[!htp]
\begin{center} 
\subfigure{
\includegraphics[width=0.95\linewidth]{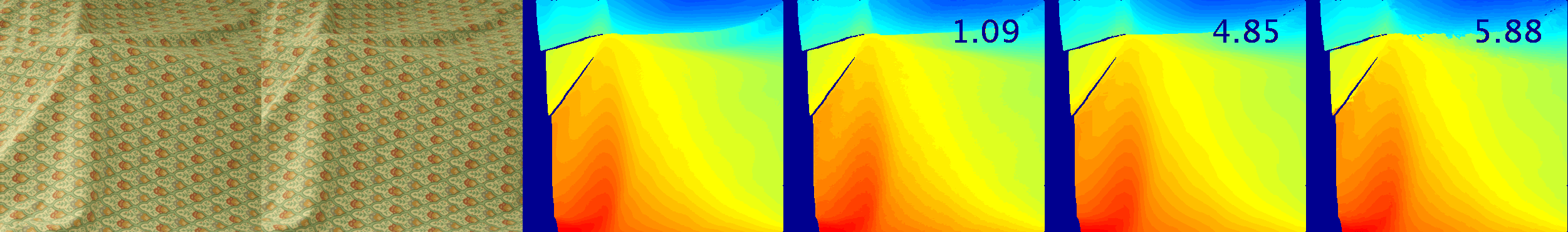} }\vspace{-0.2cm} 
\subfigure{
\includegraphics[width=0.95\linewidth]{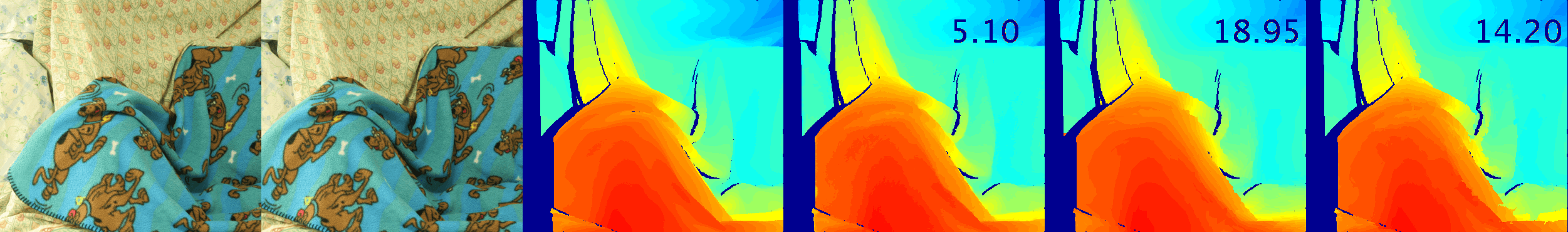} }\vspace{-0.2cm} 
\subfigure{
\includegraphics[width=0.95\linewidth]{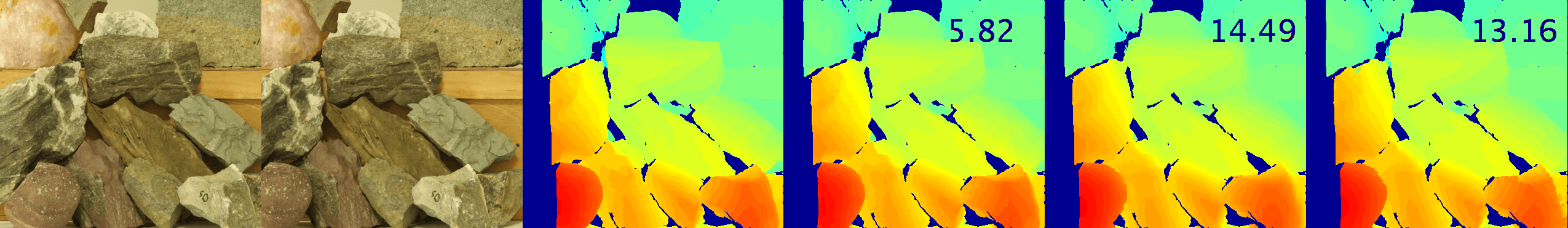} }
\caption{\label{fig:result_mb}\textbf{Results on the Middlebury stereo benchmark:} Left to right: left image, right image, ground-truth disparity map, our estimated disparity map, result of SPS-St \cite{Yamaguchi14}, and MeshStereo\cite{Zhangiccv15}. For quantitative comparison, the \emph{D1-all} error with 0.5 pixel threshold is marked on the upper right corner of all results.} 
\end{center}
\end{figure*}

\begin{figure*}[!htp]
\begin{center} 
\subfigure{
\includegraphics[width=0.47\linewidth]{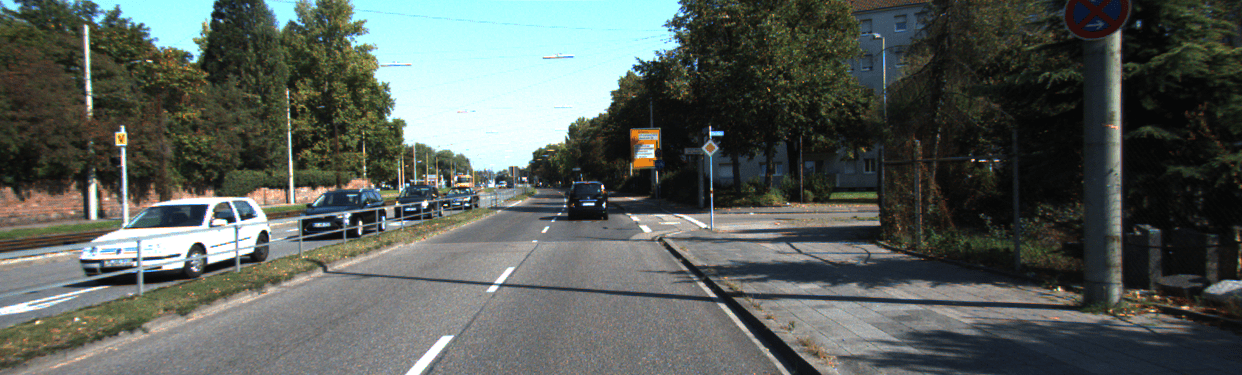} } \vspace{-0.2cm} 
\subfigure{
\includegraphics[width=0.47\linewidth]{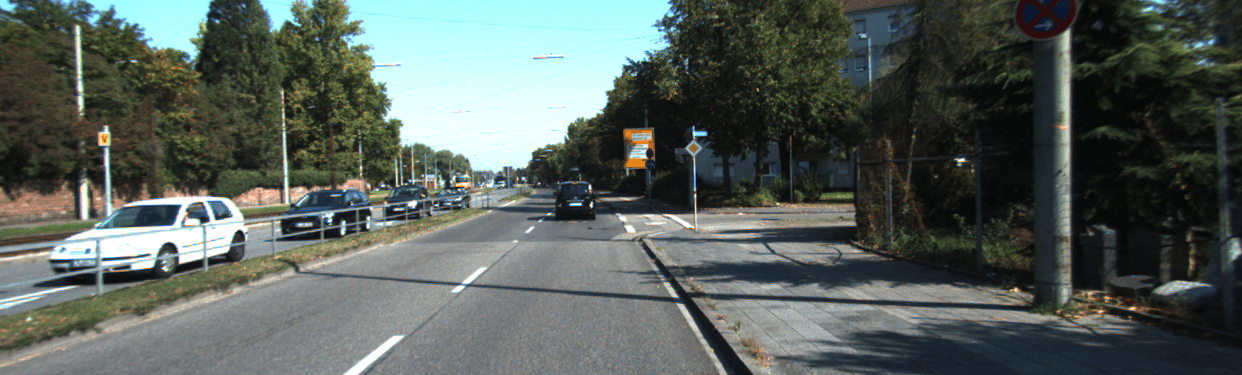} } 
\subfigure{
\includegraphics[width=0.47\linewidth]{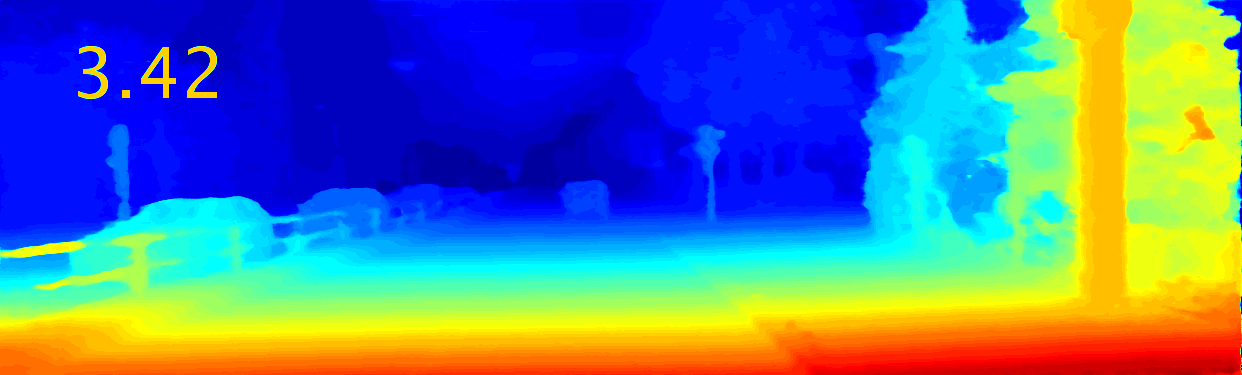} } \vspace{-0.2cm} 
\subfigure{
\includegraphics[width=0.47\linewidth]{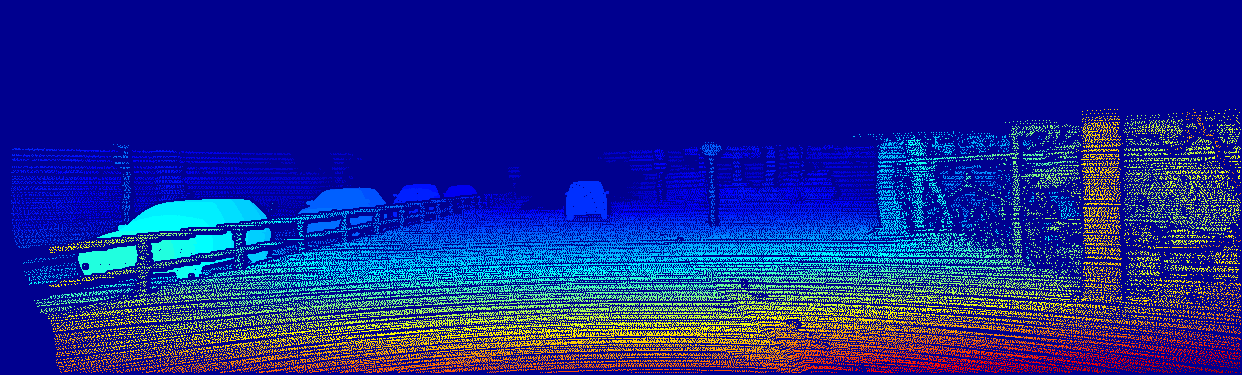} }
\subfigure{
\includegraphics[width=0.47\linewidth]{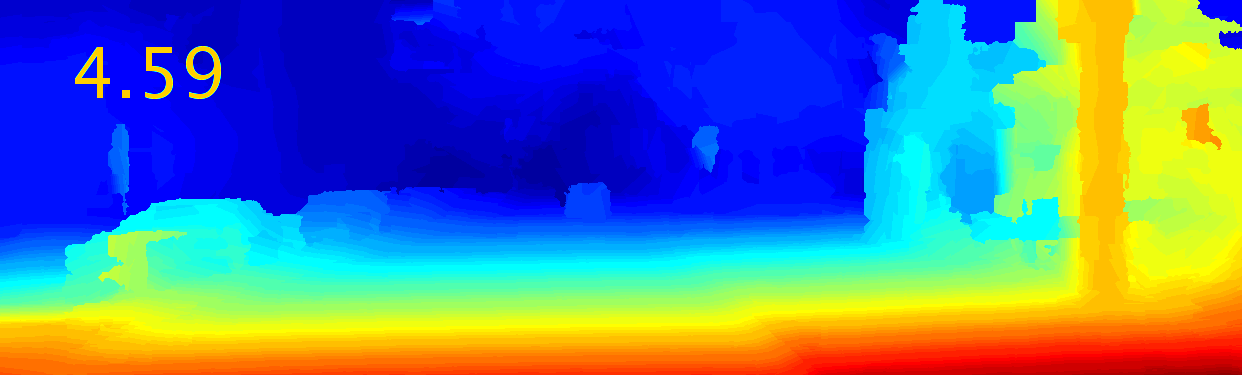} } 
\subfigure{
\includegraphics[width=0.47\linewidth]{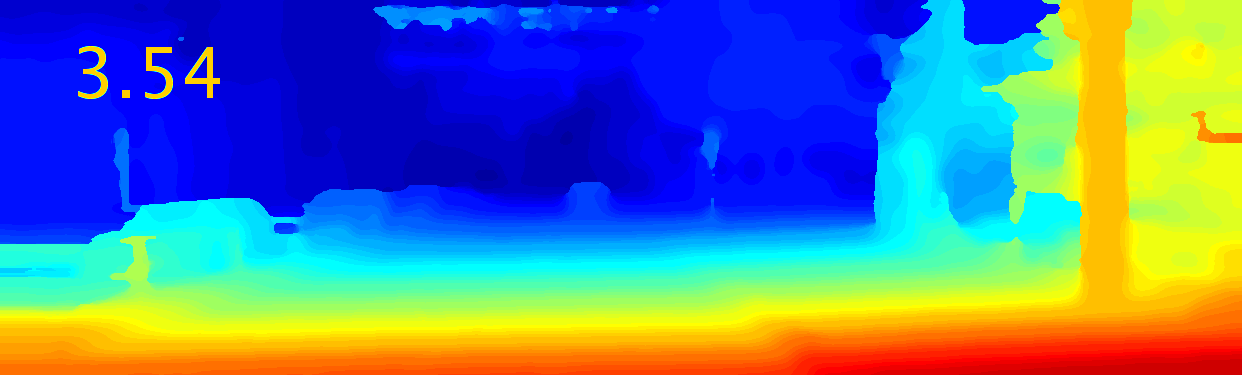} }
\caption{\label{fig:result} \textbf{Results on KITTI-2015 dataset:}  The first row shows an input stereo pair and the middle row illustrates our result and the ground truth depth. Disparity maps generated by two state-of-the-art methods are shown on the bottom row (left: SPS-St \cite{Yamaguchi14}; right: MC-CNN-acrt \cite{Zbontar2016}). The numbers shown on the recovered depth map are the quantitative comparison (\emph{D1-all} with 3 pixel threshold).}
\end{center}
\end{figure*}



Importantly, our deep stereo matching network is {\em self-adaptive}, in the sense it can adapt itself to different new scenarios, under different lighting conditions, and different camera settings. Some sample results on the KITTI dataset and on the Middlebury dataset by our method and comparison with other methods are illustrated in Fig.~\ref{fig:result_mb} and Fig.~\ref{fig:result}, where our self-supervised stereo matching method outperforms competing traditional and supervised deep learning based methods.


%

\section{Related work}

Estimating a dense depth/disparity map from a stereo image pair is a long lasting problem that has been studied for decades. Interested readers are referred to \cite{Scharstein2002}, \cite{HamzahI16} and \cite{Geiger-Survey-2017} for overviews. In this section, we provide a brief discussion on related works.

\textbf{Traditional Stereo Matching:} In general, stereo matching methods can be roughly classified as local methods and global methods. Local methods such as \cite{Scharstein2002}, SGM \cite{Hirschmuller2008} and \cite{Zhang2009} aim at finding the matching points of given points within a predefined support window. On the other hand, global methods treat disparity assignment as an optimization problem that minimize a global energy function for all disparity values. Global methods generally achieve good performance but have high computation complexities. In most cases, the resultant optimization is NP-hard. Researchers have leveraged graph cut \cite{Kolmogorov2001} or belief propagation \cite{Klaus2006} to get suboptimal results. Additionally, parametric models such slanted plane have been introduced to reduce the optimization parameters \cite{Bleyer2011, Yamaguchi14}. When ground truth depth maps are available, traditional stereo matching methods such as \cite{Zhang2007} \cite{Pal2012} and \cite{Li2008} could learn the meta parameters for Markov random field (MRF) and conditional random field (CRF) to adapt to different datasets.

\textbf{Deep Stereo Matching:} Recently, stereo matching has been greatly advanced thanks to deep convolutional neural networks (CNN). These state-of-the-art deep stereo matching models can be roughly classified into three categories: i) learn better feature correspondences \cite{Zbontar2016,Luo2016}, ii) learn better regularization \cite{Seki2017CVPR}, and iii) learn the dense disparity map in an end-to-end way \cite{Mayer2016CVPR,KendallMDHKBB17}. The first category of methods replace the handcrafted features with more distinguishable learned deep features in computing matching costs and apply non-trained traditional cost aggregation and regularization \cite{Zbontar2016,Luo2016}. The second category of methods learn the regularization and cost aggregation. Seki \etal \cite{Seki2017CVPR} learned the spatial-variant penalty-parameters of the regularization part in SGM. The last category of methods formulate stereo matching as a supervised regression or multi-class classification task and solve it in an end-to-end learning framework \cite{Mayer2016CVPR, KendallMDHKBB17}. DispNet \cite{Mayer2016CVPR} directly computes the correspondence field between stereo images, which attempts to predict the per-pixel disparity by minimizing a regression training loss. GC-Net \cite{KendallMDHKBB17} explicitly learns feature extraction, cost volume, and regularization function all in neural network. The very recent CRL (cascade residual learning) \cite{pang2017cascade} is a cascade CNN architecture composing of two stages, which follows the coarse-to-fine or residual learning principle.

\textbf{Unsupervised monocular depth learning:} Stereo matching is also closely related to monocular depth estimation, where the task is to estimate a dense disparity map from a single monocular image. Recently, novel view synthesis has been used to supervise the network learning by exploiting the availability of stereo images and image sequences \cite{garg2016unsupervised,monodepth17, zhou2017unsupervised, Xie2016}. These methods generally recast monocular depth estimation as a parametric image warping problem: instead of using ground truth dense depth as supervisors, they minimize the image reconstruction error. However, the extension from these monocular methods to stereo matching is non-trivial. When feeding the network with stereo pairs, their performances still have a large gap even compared with traditional stereo matching methods \cite{monodepth17} and will become unstable if trained for longer.


\textbf{Unsupervised learning from video} As an self-supervised learning based method, our work is also related to visual representation learning from video, where the target is to learn generic visual features from video data in an unsupervised way. Such tasks include ego-motion and depth estimation \cite{zhou2017unsupervised}, image matching \cite{Long2016}, video prediction \cite{Lecun:ICLR2016}, and video frame synthesis \cite{Liu:ICCV2017}. 


\section{Our Method}

In this section, we present our self-supervised learning based stereo matching network, which could be trained in an end-to-end way and without the need of ground truth disparity maps. We represent self-supervised stereo matching as finding the disparity map that best warp between the stereo image pair. self-supervised learning also enables the self-improving ability of our network, \ie, the network could improve the stereo matching with the evaluation of new stereo pair in an on-line way.


\subsection{Self-supervised stereo matching network}
Given a pair of rectified stereo images $I_L, I_R$, our task is to learn a function $f$ to predict the per-pixel dense disparity maps $d_L = f(I_L, I_R)$ and $d_R = f(I_R, I_L)$, namely, the disparity map for the left and right image correspondingly.  Most existing deep learning based supervised stereo matching methods minimize the discrepancy between the estimated disparity maps $d_L, d_R$ and the ground truth disparity maps $\overline{d}_L, \overline{d}_R$. However, traditional stereo matching algorithms can recover relatively good disparity maps without supervision. This motivates us to ask a natural question that whether we can learn the function $f$ without the need of dense disparity maps. We resort to the first geometric principle and express stereo matching as an image warping task, where the quality of image warping is evaluated as the reconstruction error between the observation and the reconstruction. The intuition is that if we can warp between the image pair properly, then we must have learned the dense disparity map. Specifically, given the left image $I_L$ and the disparity map for the right image $d_R = f(I_R, I_L)$, the right image $I_R$ can be generated by warping the left image with the dense disparity map,
\begin{equation}
I_R^{'}(u,v) = I_L(u+d_R(u,v),v),
\end{equation}
where $I_R^{'}$ is the warped right image. The discrepancy between the warped right image $I_R^{'}$ and the observed right image $I_R$ can work as supervisor in learning the function $f$. Symmetrically, the discrepancy between the warped left image $I_L^{'}$ and the observed left image $I_L$ provides another supervisor for $f$.

\begin{figure*}[!htp]
\begin{center} 
\includegraphics[width=1\linewidth]{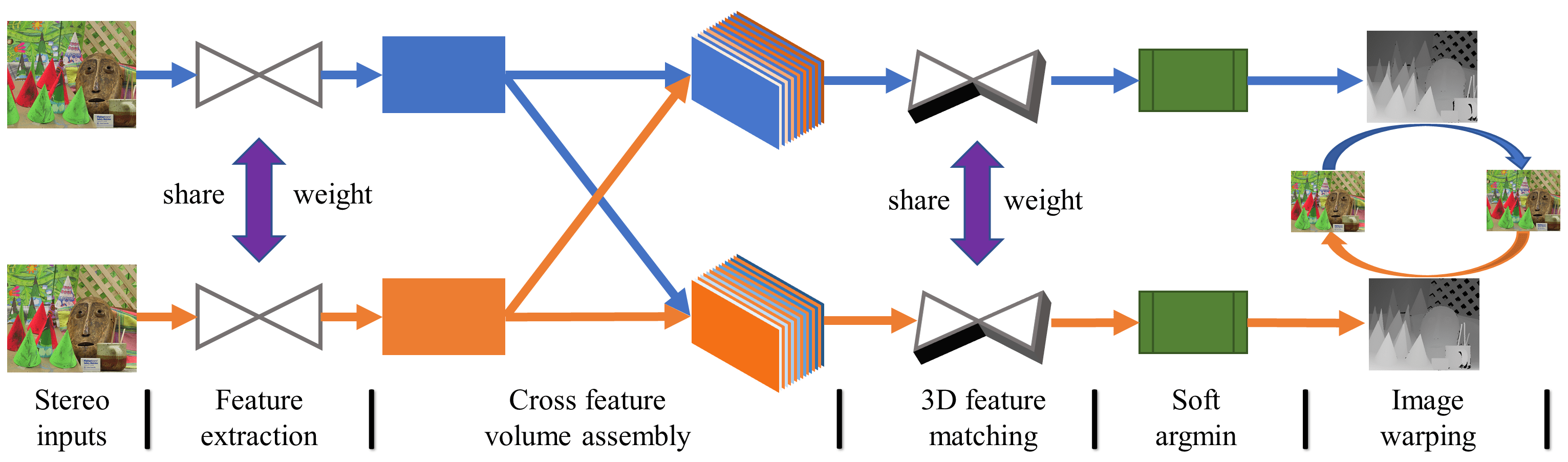} 
\caption{\label{fig:net}\textbf{Our self-supervised deep stereo matching network architecture.} Our network consists of five modules, namely, feature extraction, cross feature volume, 3D feature matching, soft-argmin, and warping loss evaluation.}
\end{center}
\end{figure*}

In this paper, we propose to learn the function $f$ by using a deep convolutional neural network in an self-supervised and end-to-end way, which basically follows the procedure in the traditional stereo matching pipeline but with a network realization. In Fig.~\ref{fig:net}, we illustrate the architecture of our self-supervised deep stereo matching network, which consists of five modules: feature extraction, feature volume generation, 3D feature matching, soft argmin and image warping. The feature extraction module consists of a series of 2D convolutions with residual connections to extract local features. These learned features from a stereo pair are assembled into two cross feature volumes. After that, feature matching (regularization) module is used to map 2D features to a higher dimensional space to make them more distinguishable. We use soft-argmin to project 3D volume to 2D. In the last module, we perform image warping to evaluate the photometric error and use it as a supervisor signal to train our network. We will discuss each module in the following subsections.





\subsubsection{Feature Extraction}

It is widely believed that feature descriptors can better capture local context, thus more robust to photometric differences (occlusion, non-lambertian lighting effects and perspective effects). In our network, instead of computing the stereo matching costs on the raw pixel intensities, we propose to use learned local features, which are also learned in self-supervised way without ground truth supervision. 

Inspired by the very recent GC-Net \cite{KendallMDHKBB17}, we design a feature extraction module with 18 convolution layers of $3\times 3$ kernels and skip connections every 3 layers. The output feature dimension is 64. We leverage symmetric feature extraction for both views, which requires the same respond for the same input. Such symmetry properties can be easily implemented in a network by sharing weights between feature extractors. We form the unary features by passing both left and right images through the feature extraction module. 



\subsubsection{Feature Volume Construction}

\begin{figure}[!htp]
\begin{center} 
\includegraphics[width=0.8\linewidth]{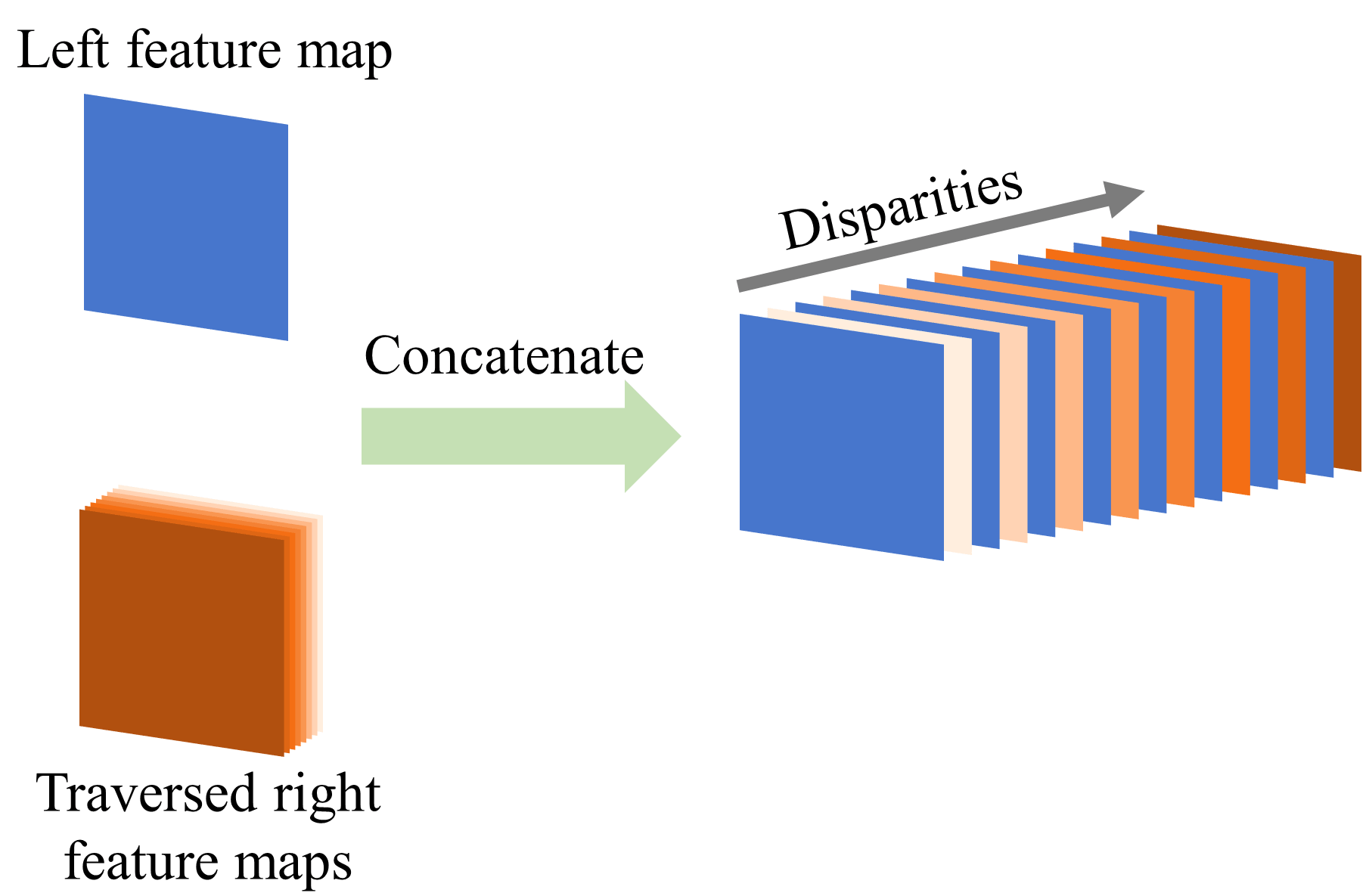}
\caption{\label{fig:vol}\textbf{Feature Volume Construction.} The cross feature volume is constructed by concatenating the learned features extracted from the left and right images correspondingly. The blue rectangle represents a feature map from the left image, the stacked orange rectangle set represents traversed right feature maps from 0 toward a preset disparity range $D$. Different intensities correspond to different level of disparity. Note that the left feature map is copied $D+1$ times to match the traversed right feature maps.}
\end{center}
\end{figure}

We use the learned features to compute stereo matching cost by constructing a feature volume, which is constructed by exhausting disparity levels in a pre-defined range. Instead of constructing a cost volume by concatenating all costs with their corresponding disparities, we concatenate the learned features from the left and right images at each disparity level and assemble a feature volume as illustrated in Fig.~\ref{fig:vol}.

Denote $f_L, f_R$ as the corresponding feature maps extracted from $I_L$ and $I_R$ by using our feature extraction module, the left-to-right feature volume at pixel position $(u,v)$ with disparity $d$ is given by:
\begin{equation}
F^{LR}(u,v,d) = f_L(u,v)\concat f_R(u-d,v),
\end{equation}
where $\concat$ denotes the vector concatenation operation. Correspondingly, the right-to-left feature volume is
\begin{equation}
F^{RL}(u,v,d) = f_R(u,v)\concat f_L(u+d,v).
\end{equation}
In this way, we reach a feature volume with dimension $height\times width \times (max~disparity+1) \times feature~dimension$ for the left-to-right and right-to-left feature volume correspondingly.


\subsubsection{3D Feature Matching with Regularization}
\begin{figure*}[!htp]
\begin{center} 
\includegraphics[width=\textwidth]{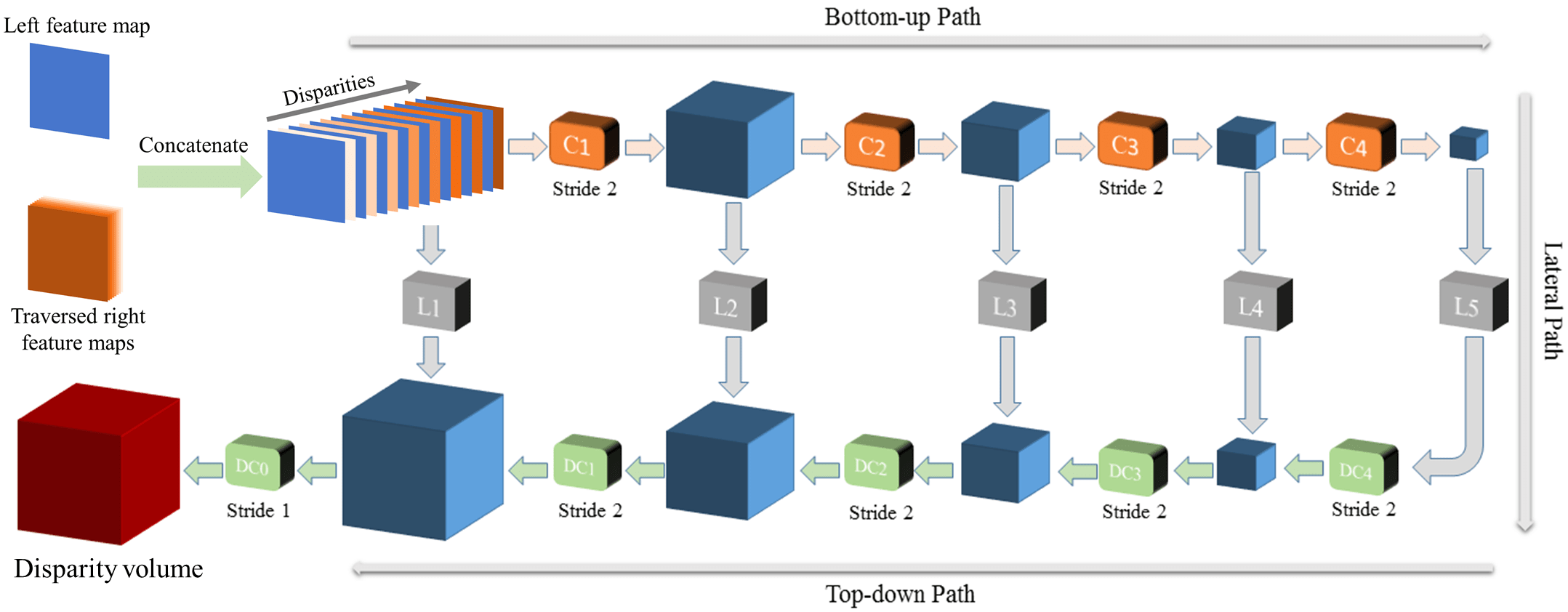}
\caption{\label{fig:resTDM}\textbf{Diagram of our res-TDM module for 3D feature matching with learned regularization.}  It takes cross feature volume as an input, and is followed by a series of 3D convolution and deconvolution. $C_i$ denotes the 3D convolution layer, $R_i$ is the residual module that connects low-level features to the top-down pathway. $DC_i$ is the 3D deconvolution layer for upsampling. The output of this module is a 3D disparity volume of dimension $H\times W \times (D+1)$.}
\end{center}
\end{figure*}

With the assembled feature volume, we would like to learn the matching cost at each candidate disparity not only with the unary term but also with the regularization from local context. As our feature volume owns 4 dimensions, namely height, width, disparity range and feature dimension. We propose to use 3D convolutions rather than 2D convolutions, which is able to exploit the correlation in height, width and disparity direction. We present a Residually connected Top-Down Module (Res-TDM) for extracting better features with the mixture of disparity and spatial location. A nutshell of our Res-TDM is shown in Fig.~\ref{fig:resTDM}. In the Bottom-up phase, the 3D volume ($H\times W\times (D+1)\times 2F$) passes through a series of 3D convolutional layers ($C_i$) with the same kernel size $3\times 3 \times 3$ and a stride 2 until achieving an encoded feature volume with dimension $(1/16) H\times (1/16)W\times (1/16)(D+1)\times F$, where $H,W,D,F$ represent the height, width, disparity range, and feature respectively. In the Top-down phase, a mirrored process scales up the encoded feature volume back to the original dimension by swapping the 3D convolution with 3D deconvolution. For each scale, we apply our Res-TDM with a residual module $R_i$. Each $R_i$ consists of two 3D convolution layers with the same kernel size $3\times 3 \times 3$ and stride 1.

\subsubsection{Soft Argmin}
The output of our Res-TDM module is a 3D volume with regularized features. However, for image warping, a 2D disparity map is needed. We naturally embed our feature matching step into this 3D to 2D process. During this step, we shrink the disparity dimension by selecting the disparity with minimal distance between left and right features in a soft-argmin way. Similar to the GC-Net \cite{KendallMDHKBB17}, we perform a soft argmin operation over the disparity dimension to project the 3D volume to 2D. The soft argmin operation is defined as:
\begin{equation}
\argmin \sum_{d=0}^D d \times \sigma(-c_d),
\end{equation}
where $c$ is the predicted cost (similarity at disparity $d$) and $\sigma(\cdot)$ represents the softmax operation.

\subsubsection{Loss Function}




Under our self-supervised learning formulation for stereo matching, the quality of disparity map estimation is evaluated as the image reconstruction error. Our loss function for learning disparity map is defined as:
\begin{equation}
\begin{split}
\mathcal{L} = & \omega_p (\mathcal{L}_u^l+\mathcal{L}_u^r) + \omega_s (\mathcal{L}_s^l+\mathcal{L}_s^r) \\ & + \omega_c (\mathcal{L}_c^l+\mathcal{L}_c^r) + \omega_m (\mathcal{L}_m^l+\mathcal{L}_m^r),
\end{split}
\end{equation}
where $\mathcal{L}_u^l, \mathcal{L}_u^r$ denote the unary term, $\mathcal{L}_s^l, \mathcal{L}_s^r$ express the disparity field regularization term, $\mathcal{L}_c^l, \mathcal{L}_c^r$ denote the consistency constraint defined between stereo image pair and corresponding disparity maps, $\mathcal{L}_m^l, \mathcal{L}_m^r$ express the maximize depth heuristic (MDH). 

\noindent
\textbf{Unary term.} As a unary term, we would like to minimize the discrepancy between the observation and the reconstruction. It can be done by forming a loss by simply computing the $L_1$ distance between images themselves and the image gradients. Furthermore, in order to improve the robustness against illuminations, we add a structure similarity term SSIM. Therefore, our photometric based unary loss $\mathcal{L}^l_u$ is derived as:
\begin{equation}
\begin{split}
\mathcal{L}_u^l (I_L, I_L^{'}) = & \frac{1}{N}\sum\lambda_1\frac{1-\mathcal{S}(I_L, I_L^{'})}{2} \\ + & \lambda_2\left| I_L- I_L^{'}\right| + \lambda_3\left| \nabla I_L - \nabla I_L^{'}\right|,
\end{split}
\end{equation}
where $N$ is the total number of pixels and $I_L^{'}$ is the reconstructed left image. SSIM $\mathcal{S}(\cdot)$ \cite{SSIM2004} measures the structural similarity between image patches. $\lambda_1, \lambda_2, \lambda_3$ balance between structural similarity, image appearance difference and image gradient difference. We set $\lambda_1 = 0.80,\lambda_2 = 0.15,\lambda_3= 0.15$ through out our experiments. According to \cite{monodepth17}, $I_L^{'}$ can be fully differentially reconstructed from the right image $I_R$ and the right disparity map $d_R$ by bilinear sampling \cite{STN2015}.

\noindent
\textbf{Regularization term.} For regularization term, we assume the desired disparity map should be locally smooth. we leverage the Total Generalized Variation (TGV) for better subpixel level accuracy than Total Variation (TV). We also weight this term with image's second order gradients. Specifically, our smoothness based regularization for disparity field is defined as:
\begin{equation}
\mathcal{L}_s^l = \frac{1}{N}\sum\left| \nabla^2_u d_L\right| e^{-\left| \nabla^2_u I_L\right|}+ \left| \nabla^2_v d_L\right| e^{-\left| \nabla^2_v I_L\right|},
\end{equation}
where $\nabla$ denotes the gradient operator.

\begin{figure}[!htp]
\begin{center} 
\includegraphics[width=1\linewidth]{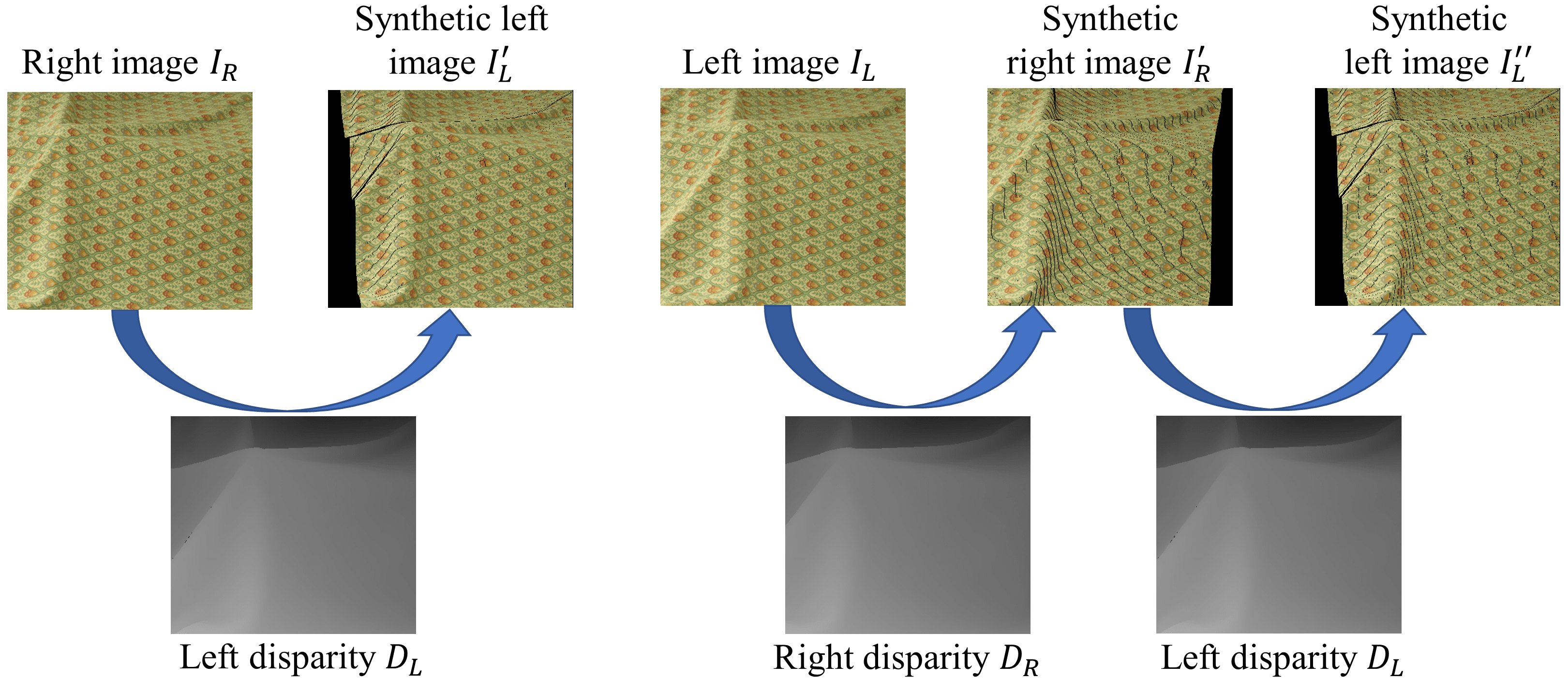}
\caption{\label{fig:lr}\textbf{Loop consistency constraint in stereo matching.} The three versions of the left image provide two constraints in regularizing the disparity maps, \ie, $I_L = I^{'}_L$, and $I_L = I^{''}_L$. The same constraint could also be derived for the right image.}
\end{center}
\end{figure}

\noindent
\textbf{Consistency term:} Besides the above regularization term defined for each disparity map separately, we further apply a new loop consistency term in our model by considering the consistency between the disparity maps for the left and right images. An illustration of our loop consistency constraint is illustrated in Fig.~\ref{fig:lr}. Given a left image, we can synthesize its two versions by using the disparity maps and the images. The first synthesized left image $I^{'}_L$ is generated by warping the right image to the left image coordinate with the disparity map defined on the right image. The second synthesized left image $I^{''}_L$ is generated by warping the left image to the right view and warping back to the left image coordinate by using $d_L$ and $d_R$. The three versions of the left image provide two constraints in regularizing the disparity maps, \ie, $I_L = I^{'}_L$, and $I^{'}_L = I^{''}_L$. The same constraints could also be derived for the right image. Thus our loop consistency loss $\mathcal{L}_{c}^L$ is defined as:
\begin{equation}
\mathcal{L}_c^L = |I_L-I^{''}_L|.
\end{equation}
Note that the left-right consistency term proposed in Godard \etal \cite{monodepth17} is a linear approximation of our loop constraint.

It is worth noting that this loop consistency plays a key role in tightly coupling our symmetric network. Without this loss, our symmetric network can always be decoupled into two networks equivalently. The loop consistency enables our network to make the full benefit of the symmetric structure. 





\noindent
\textbf{Maximum-Depth Heuristic} In real world scenarios, there may be multiple warping functions that achieve similar warping loss, especially for the textureless areas. To further provide strong regularization in handling textureless regions, we propose to leverage the Maximum-Depth Heuristic (MDH) \cite{Perriollat:BMVC08} in our model, which maximizes the sum of all depths or minimizes the sum of all the disparities. Therefore, we define a MDH loss as:
\begin{equation}
\mathcal{L}_m^L = \frac{1}{N}\sum\left| d^L\right|. 
\end{equation}


\section{Self-improving Ability}
\begin{figure*}[!htp]
\begin{center} 
\subfigure{
\includegraphics[width=0.48\linewidth]{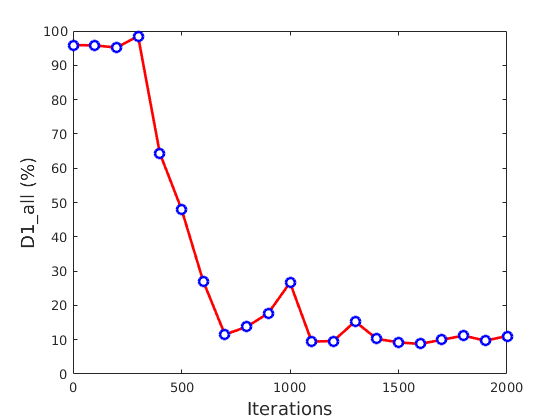} } 
\subfigure{
\includegraphics[width=0.48\linewidth]{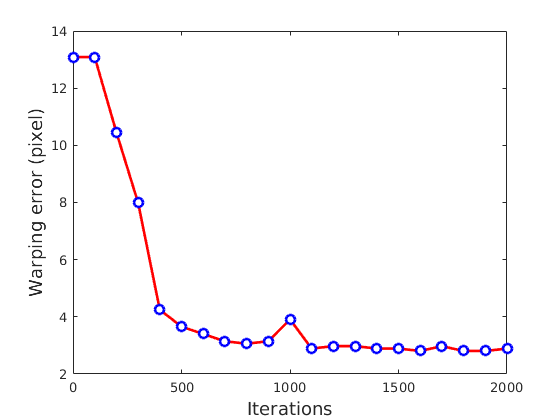} }
\caption{\label{fig:evo}\textbf{Self-improving Curves.} The left figure shows that our network can achieve reasonable results within 1500 iterations. The right one shows the warping error along with training iterations. They both show a similar trend in learning process.}
\end{center}
\end{figure*}

Our network can be applied in two different modes. One is the traditional mode, where the training stage and testing stage are clearly separate, and during testing stage the network's all parameters (expect for input) are frozen.  The other mode is what we call the "self-improving" mode where the network is allowed to continuously fine-tune its parameters while testing on new stereo images in a new environment.  This latter mode effectively gives our network the ability to adapt itself to new never-seen-before scenarios. In other words, it can be "automatically" generalize to unseen images.  This is possible because we do not require ground-truth depth-maps during training; instead, input stereo pairs serve as self-supervision signals, and the network is able to iteratively self-improve automatically. 

\begin{figure*}[!htp]
\begin{center} 
\subfigure{
\includegraphics[width=0.4\linewidth]{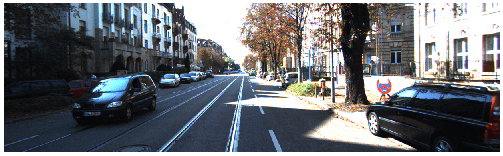} } \vspace{-.2cm} 
\subfigure{
\includegraphics[width=0.4\linewidth]{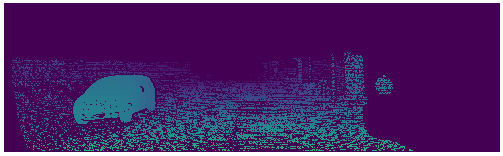} } 
\rot{\scriptsize   Ground Truth}
\subfigure{
\includegraphics[width=0.4\linewidth]{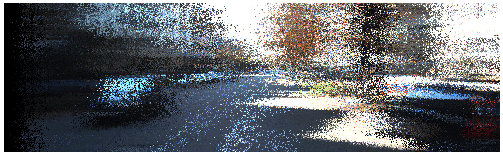} } \vspace{-.2cm} 
\subfigure{
\includegraphics[width=0.4\linewidth]{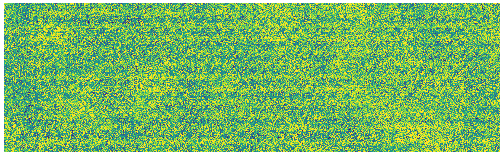} }
\rot{\scriptsize Iteration 0}
\subfigure{
\includegraphics[width=0.4\linewidth]{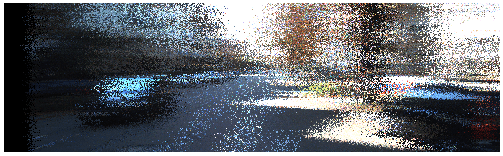} } \vspace{-.2cm} 
\subfigure{
\includegraphics[width=0.4\linewidth]{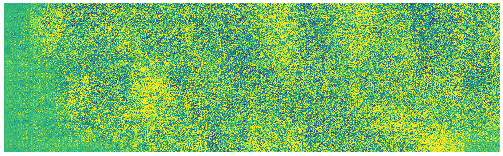} }
\rot{\scriptsize Iteration 100}
\subfigure{
\includegraphics[width=0.4\linewidth]{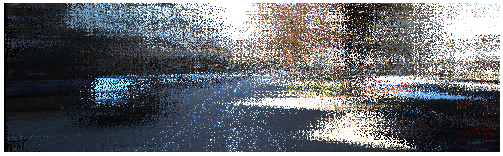} } \vspace{-.2cm} 
\subfigure{
\includegraphics[width=0.4\linewidth]{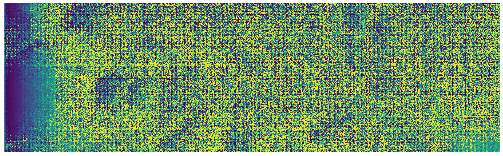} }
\rot{\scriptsize Iteration 200}
\subfigure{
\includegraphics[width=0.4\linewidth]{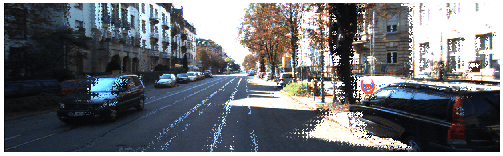} } \vspace{-.2cm} 
\subfigure{
\includegraphics[width=0.4\linewidth]{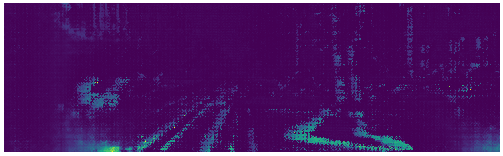} }
\rot{\scriptsize Iteration 300}
\subfigure{
\includegraphics[width=0.4\linewidth]{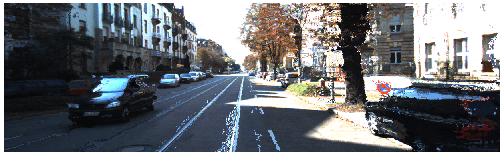} } \vspace{-.2cm}
\subfigure{
\includegraphics[width=0.4\linewidth]{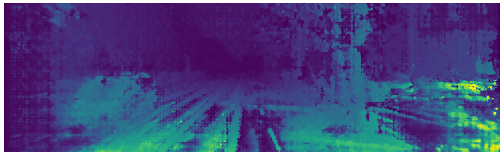} }
\rot{\scriptsize Iteration 400}
\subfigure{
\includegraphics[width=0.4\linewidth]{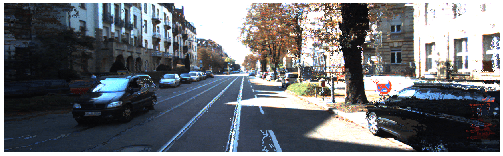} } \vspace{-.2cm}
\subfigure{
\includegraphics[width=0.4\linewidth]{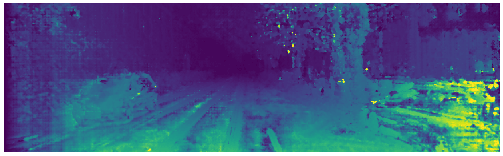} }
\rot{\scriptsize Iteration 500}
\subfigure{
\includegraphics[width=0.4\linewidth]{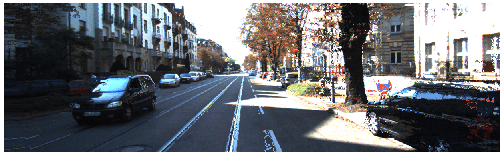} } \vspace{-.2cm}
\subfigure{
\includegraphics[width=0.4\linewidth]{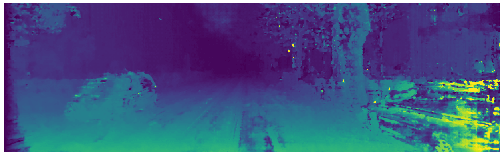} }
\rot{\scriptsize Iteration 600}
\subfigure{
\includegraphics[width=0.4\linewidth]{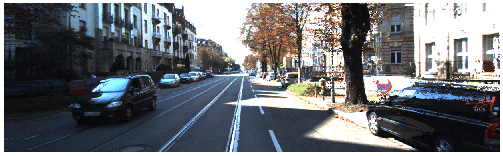} } 
\subfigure{
\includegraphics[width=0.4\linewidth]{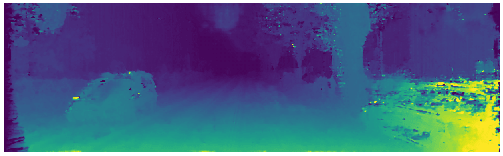} }
\rot{\scriptsize Iteration 700}
\caption{\label{fig:evoimg}\textbf{An example self-improving curve.} The left image and ground truth disparity map are on the top, followed by the inter-media results obtained after every 100 iterations.}
\end{center}
\end{figure*}

We validate this claim by testing it in numerous experiments on different types of scenes, both indoor and outdoor, and with different network initial states.  One of the tests is an extreme case where we start the learning process purely from scratch, \ie, using random network initialization, and we want to see how quickly the network is able to predict accurate depth-maps through unsupervised self-learning.  Specifically, we randomly initialize our network and then continuously feed it with random stereo image pairs, e.g. using KITTI raw dataset.  The performance of the network is then evaluated using KITTI-2015 training dataset.  Note, the evaluation signals do not feedback to the network; in other words the network is only learning blindly, and we want to find out whether or not its performance would improve. We use two quantitative metrics to measure the performance. One is the ``D1\_all", used by KITTI benchmark and the other is the ``image warping error".  Fig.~\ref{fig:evo} shows the learning curve.  Moreover, in Fig.\ref{fig:evoimg}, we show that the intermediate performance as a function of iteration time.  It is clear that, even starting from a random initialization,  after about 1000--1500 iterations our network was able to predict good depth-maps, and its performance can further improve after seeing more stereo images.

We further analyze the \emph{self-improving} ability of our network by evaluating its performance across two very different datasets: KITTI and Middlebury. We trained our network model on the KITTI raw dataset and tested the model on the Middlebury dataset. Given this baseline network model, we updated it with the new dataset. Table~\ref{tab:mb_evo} compares the improvement between the pre-trained model and the results after 100 iterations. We could observe that all the results have been greatly improved by on-line tuning, namely, the error metric decreases from $21.17\%$ to $13.67\%$ for $0.5$ threshold and from $10.80\%$ to $6.07\%$ for $1$ threshold on average. It implies that our network indeed owns an ability to improve itself by seeing more imageries.

\begin{table*}[!htp]
\centering
\scriptsize
\tabcolsep=0.126cm
\begin{tabularx}{\textwidth}{c|c|cccccccccccccccccc|c}
Model                       & \rot{Threshold} & \rot{Venus} & \rot{Dolls} & \rot{Laundry} & \rot{Moebius} & \rot{Reindeer} &\rot{Aloe} &\rot{Baby1} &\rot{Baby2} &\rot{Baby3} &\rot{Cloth1} &\rot{Cloth2} &\rot{Cloth3} &\rot{Cloth4} &\rot{Rocks1} &\rot{Rocks2} &\rot{Tsukuba} & \rot{ConesH} & \rot{TeddyH}  & Mean \\  \hline
\multirow{2}{*}{Pretrained on kitti}      & 0.5       & 15.30 & 26.43 & 36.64 & 27.57 & 27.81 & 15.35 & 19.15 & 23.34 & 31.97 & 1.66 & 12.34 & 8.30 & 9.06 & 16.70 & 8.80 & 41.57 & 26.70 & 32.43 & 21.17\\ \cline{2-21} 
                             & 1         & 8.12 & 12.65 & 25.16 & 18.50 & 17.34 & 8.62 & 10.10 & 9.56 & 17.66 & 0.80 & 5.21 & 4.17 & 5.67 & 5.38 & 4.70 & 14.65 & 9.47 & 16.59 & 10.80   \\ \hline
\multirow{2}{*}{On-line tuned}     & 0.5       & 7.27 & 17.68 & 25.79 & 18.48 & 16.46 & 10.87 & 9.07 & 13.49 & 11.29 & 1.09 & 6.90 & 5.10 & 5.30 & 12.69 & 5.82 & 37.16 & 16.85 & 24.76 & 13.67\\ \cline{2-21} 
                             & 1         & 2.86 & 7.58 & 15.93 & 12.27 & 9.30 & 5.67 & 4.32 & 4.00 & 6.16 & 0.42 & 2.64 & 2.52 & 3.08 & 2.94 & 2.69 & 11.90 & 5.10 & 9.90 & 6.07\\ \hline
\end{tabularx}
\caption{\label{tab:mb_evo}\textbf{Self-improving on the Middlebury stereo dataset.} We compare the performance between the pre-trained model on KITTI and the one that on-line tuned for 100 iterations.}
\end{table*}

\section{Experiment}
In this section, we compare the performance of our method with state-of-the-art stereo matching methods. Our network is trained end-to-end on rectified stereo image pair in an self-supervised way without any post-processing or requiring any ground truth depth maps. We report qualitative and quantitative results on three datasets: KITTI stereo 2012 \cite{Geiger2012CVPR}, KITTI stereo 2015 \cite{Menze2015CVPR}, Middlebury stereo \cite{Scharstein2003, Scharstein07, HirschmullerS07}.

\subsection{Implementation Details}
Our network is implemented in TensorFlow \cite{tensorflow}, which could provide a reasonable result within 1500 iterations when trained from scratch. Since there is no clear distinction between training phase and testing phase for our network (the only difference is that whether the network parameters need to update). In the inference period (without updating parameters), it takes about 0.8 second to process a stereo pair with resolution $384\times 1280$, including data loading and transferring times. Such processing time will increase to 1.6 seconds when on-line tuning is performed.


All models are optimized end-to-end with RMSProp \cite{Tieleman2012} with an initial learning rate of $1\times 10^{-3}$, $1\times 10^{-4}$ after 5000 iterations. The input images are randomly cropped from a pair of normalized stereo images with pixel intensities ranging from 0 to 1. No data augmentation has been used in our experiments. Due to the hardware limit, we set the batch size to 1, input resolution as $256\times 512$ during training. For 3D feature matching, we set the disparity range to $160$. For weighting different loss components, when training from scratch, $\omega_s$ need to be set equal or less than $0.001$ in order to avoid a trivial solution: all pixels have been assigned by the maximum disparity. However, $\omega_s$ can be increased to $0.1$ when the network is converged. We fix $\omega_c = 1, \omega_m = 0.001$ for all experiments.

\subsection{KITTI}
We trained our network on KITTI raw data that consists of 42,382 rectified stereo pairs from 61 scenes with a typical image size $1242\times 375$. Note that there is no split of training or testing as our network is totally self-supervised. 

Evaluation is done on KITTI-2012 \cite{Geiger2012CVPR} and KITTI-2015 \cite{Menze2015CVPR} stereo datasets. KITTI-2012 consists of 194 training pairs and 195 testing pairs while KITTI-2015 contains 200 stereo pairs for training and 200 stereo pairs for testing. In Table~\ref{tab:kitti12} and Table~\ref{tab:kitti15}, we evaluate the performance of our model on KITTI-2012 (2 pixels threshold) and KITTI-2015 testing subsets respectively. The ground truth disparities for testing dataset are withheld for evaluation. 

\begin{figure*}[!htp]
\begin{center} 
\subfigure{
\includegraphics[width=0.32\linewidth]{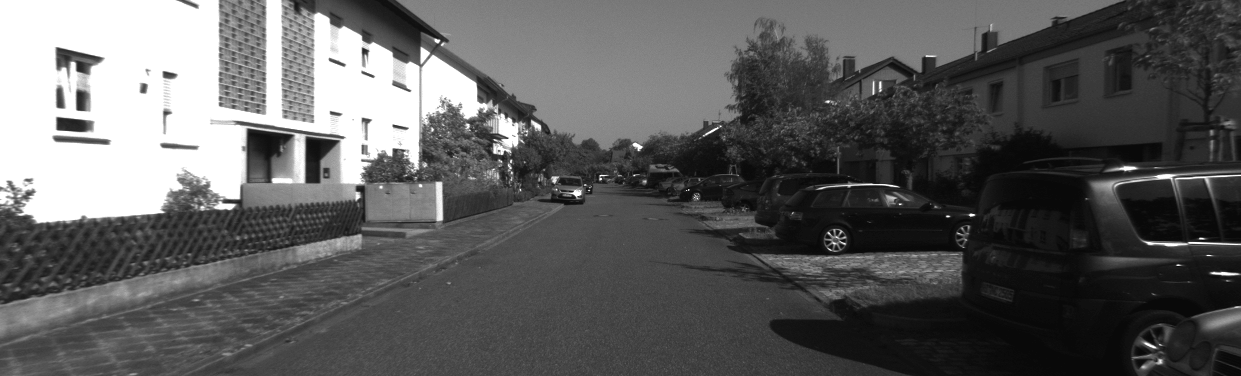} } 
\subfigure{
\includegraphics[width=0.32\linewidth]{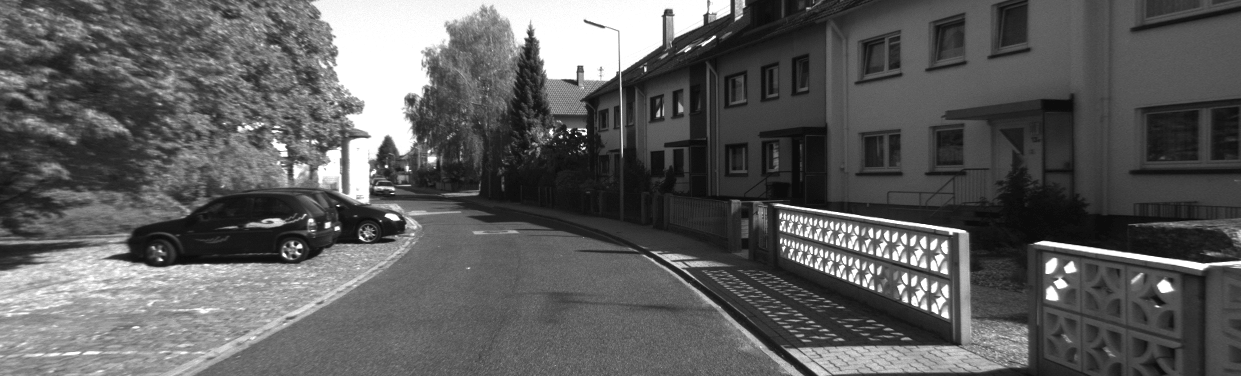} } 
\subfigure{
\includegraphics[width=0.32\linewidth]{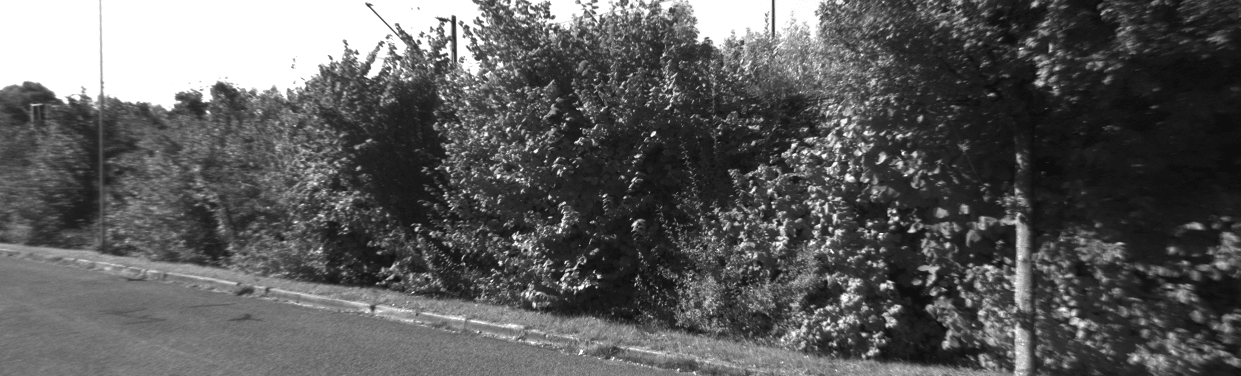} } 
\subfigure{
\includegraphics[width=0.32\linewidth]{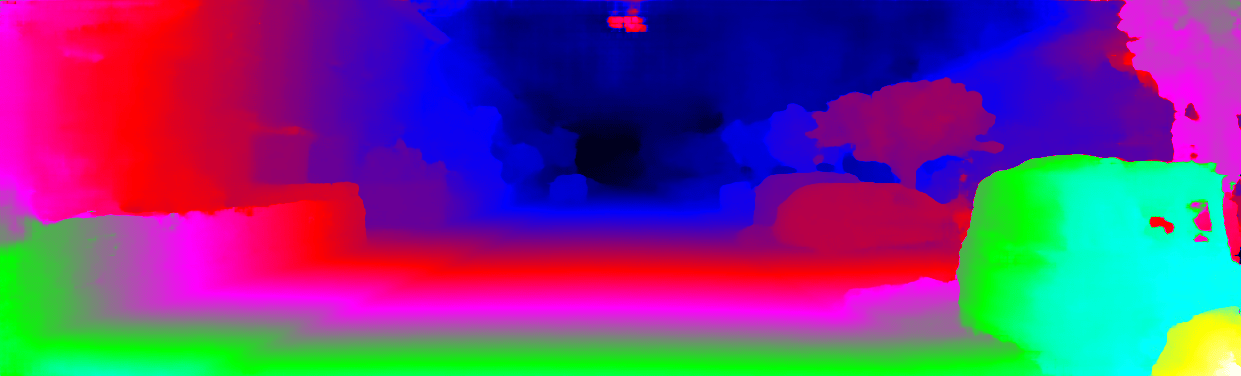} } 
\subfigure{
\includegraphics[width=0.32\linewidth]{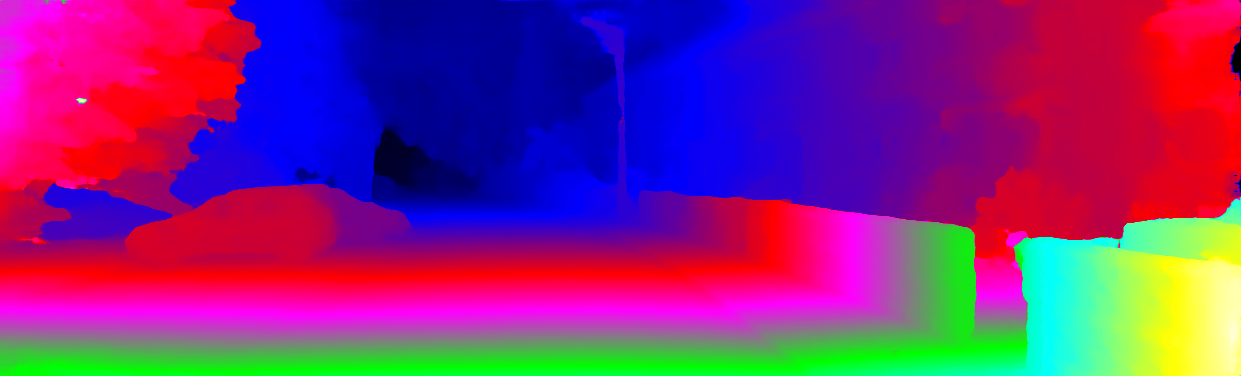} } 
\subfigure{
\includegraphics[width=0.32\linewidth]{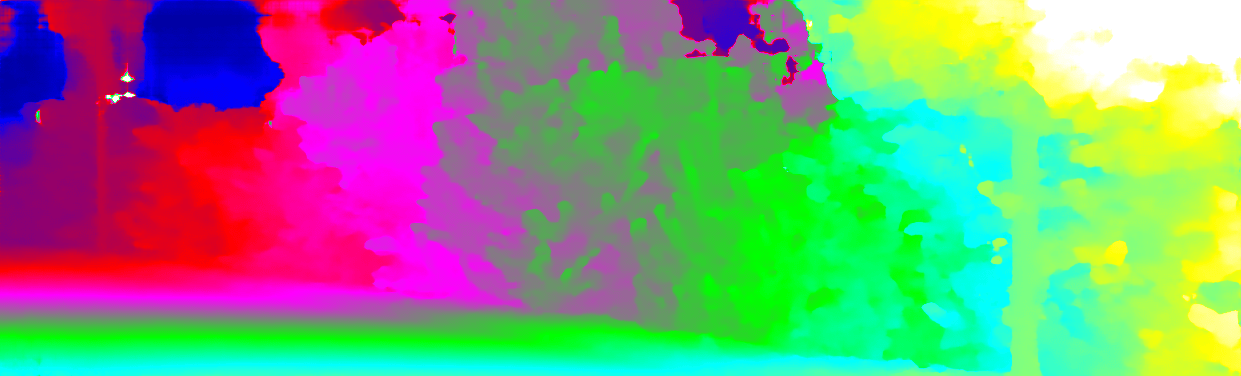} } 
\subfigure{
\includegraphics[width=0.32\linewidth]{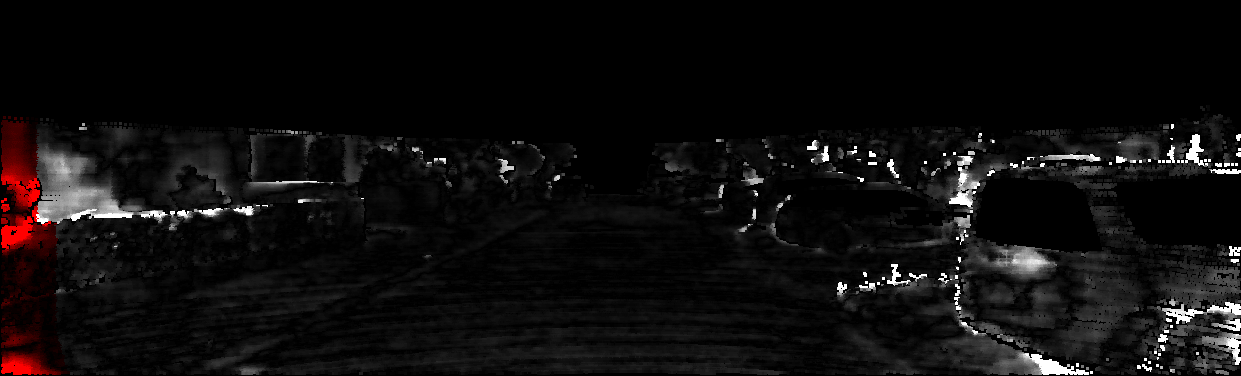} } 
\subfigure{
\includegraphics[width=0.32\linewidth]{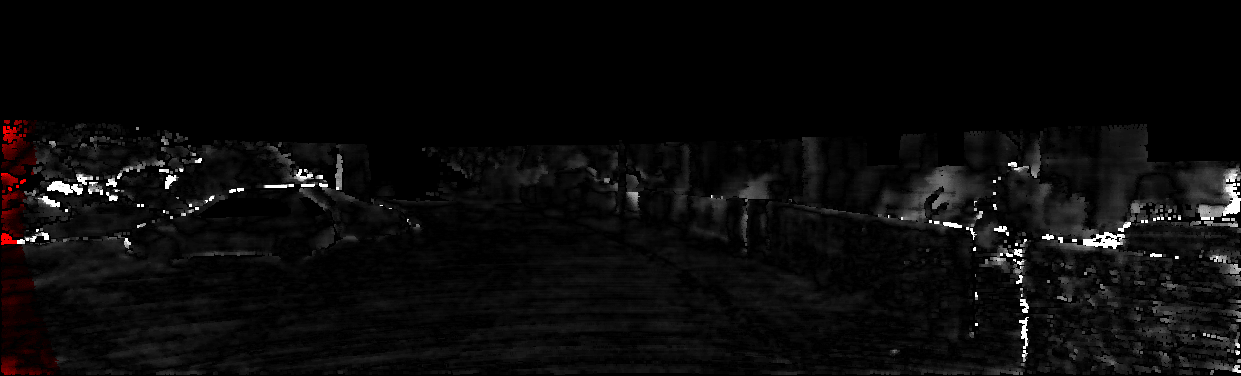} } 
\subfigure{
\includegraphics[width=0.32\linewidth]{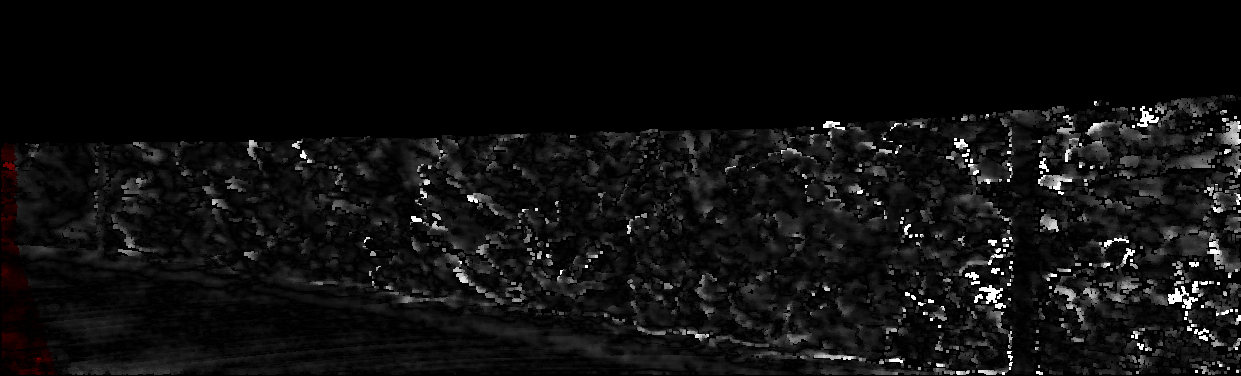} } 
\subfigure{
\includegraphics[width=0.32\linewidth]{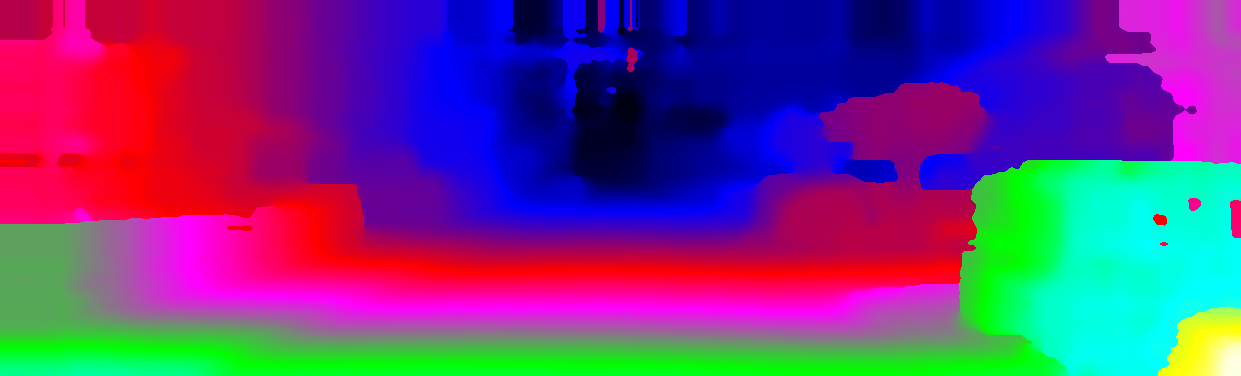} } 
\subfigure{
\includegraphics[width=0.32\linewidth]{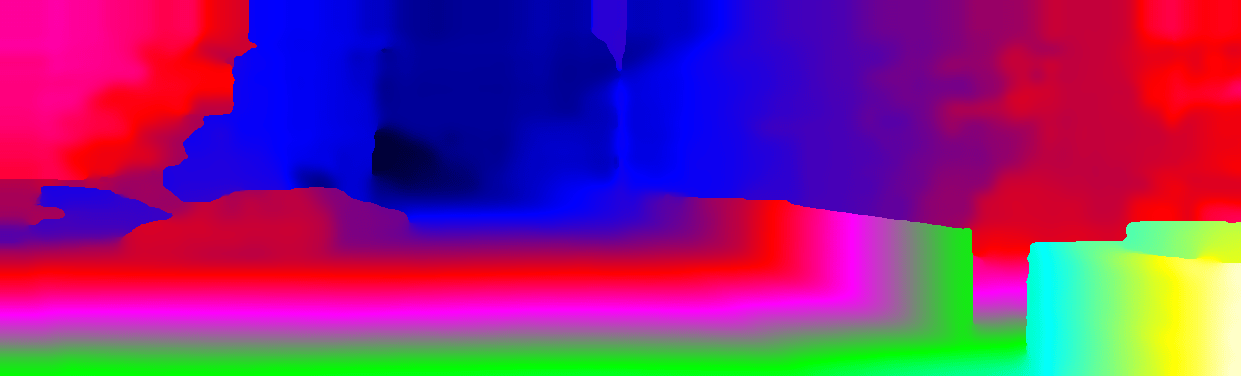} } 
\subfigure{
\includegraphics[width=0.32\linewidth]{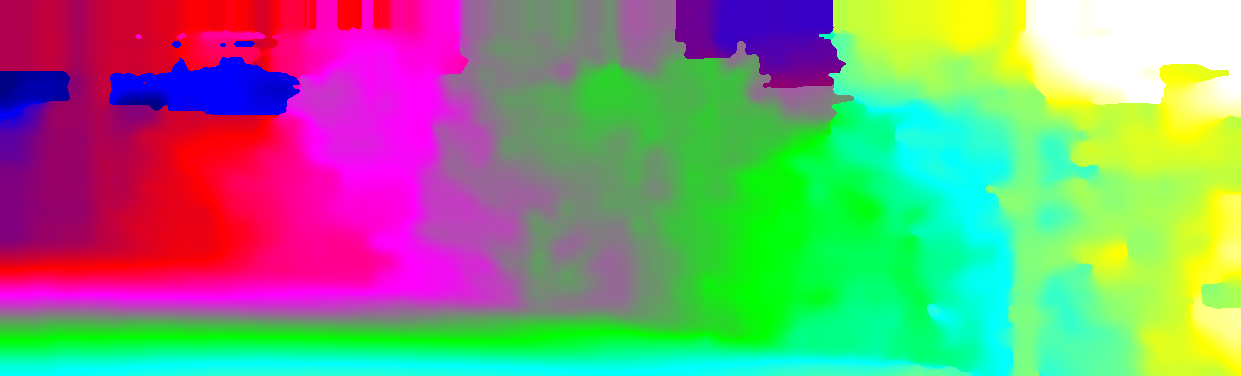} } 
\subfigure{
\includegraphics[width=0.32\linewidth]{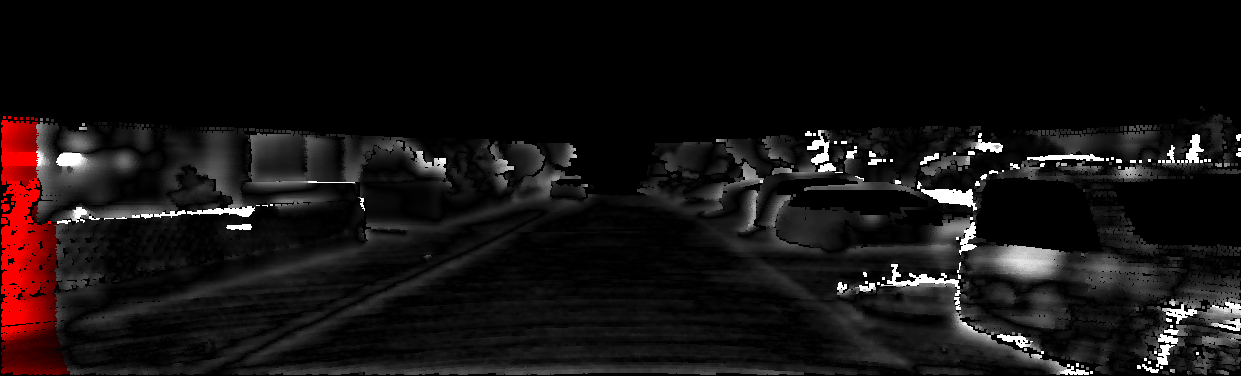} } 
\subfigure{
\includegraphics[width=0.32\linewidth]{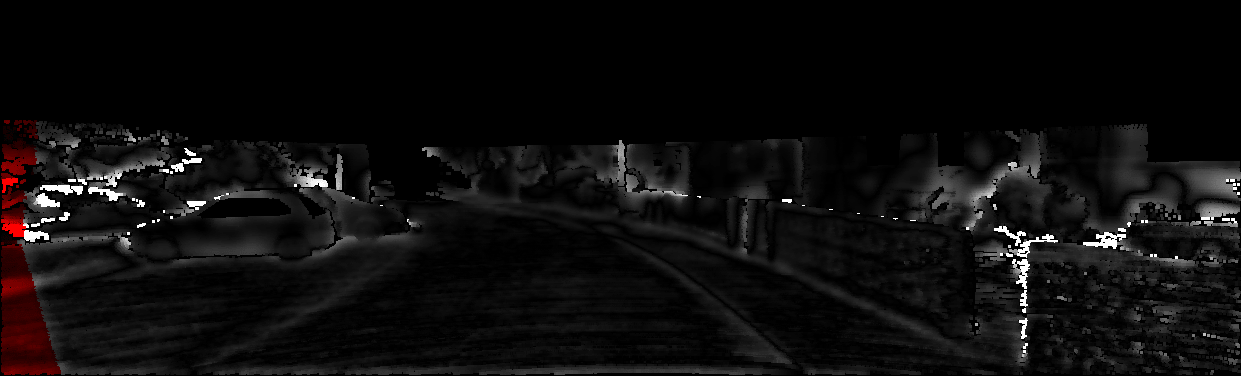} } 
\subfigure{
\includegraphics[width=0.32\linewidth]{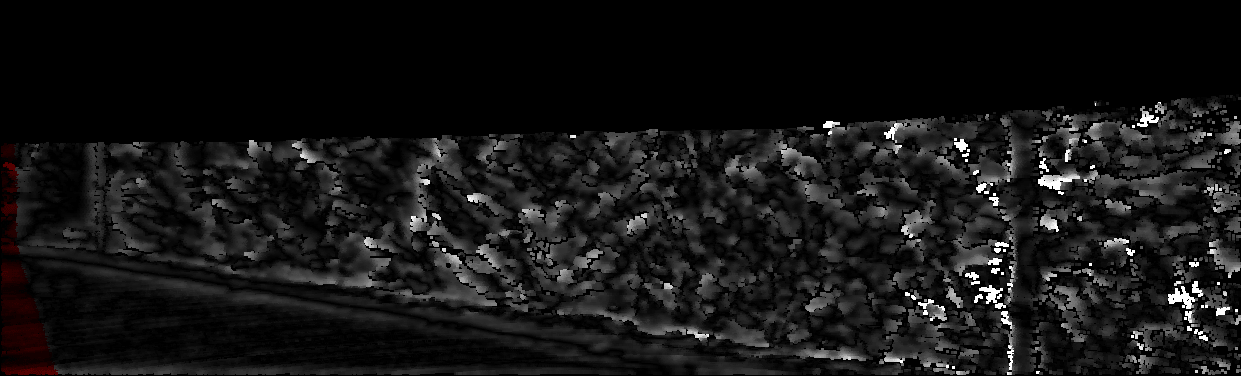} } 
\caption{\label{fig:kitti12}\textbf{Qualitative evaluations on KITTI-2012:} Top to bottom: left image, our result, our error map, result of MC-CNN-acrt \cite{Zbontar2016} and its error map.}
\end{center}
\end{figure*}
\begin{figure*}[!htp]
\begin{center} 
\subfigure{
\includegraphics[width=0.32\linewidth]{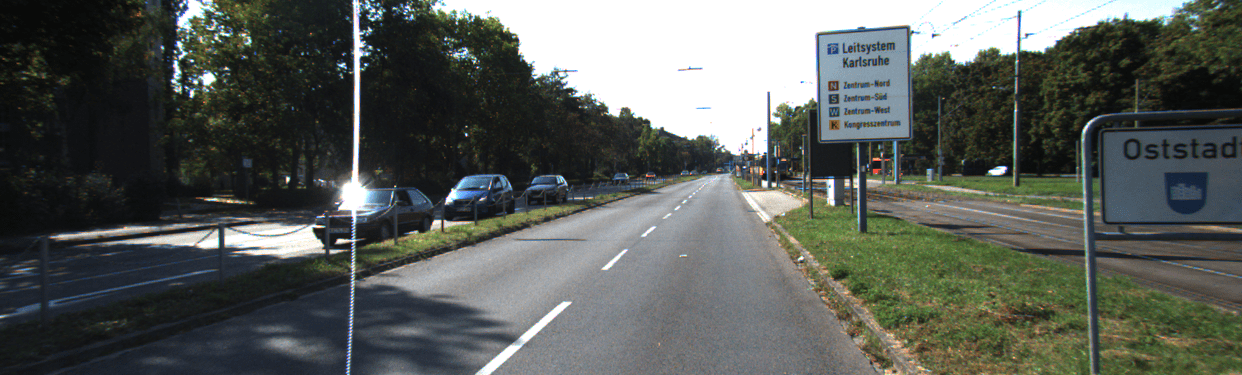} } 
\subfigure{
\includegraphics[width=0.32\linewidth]{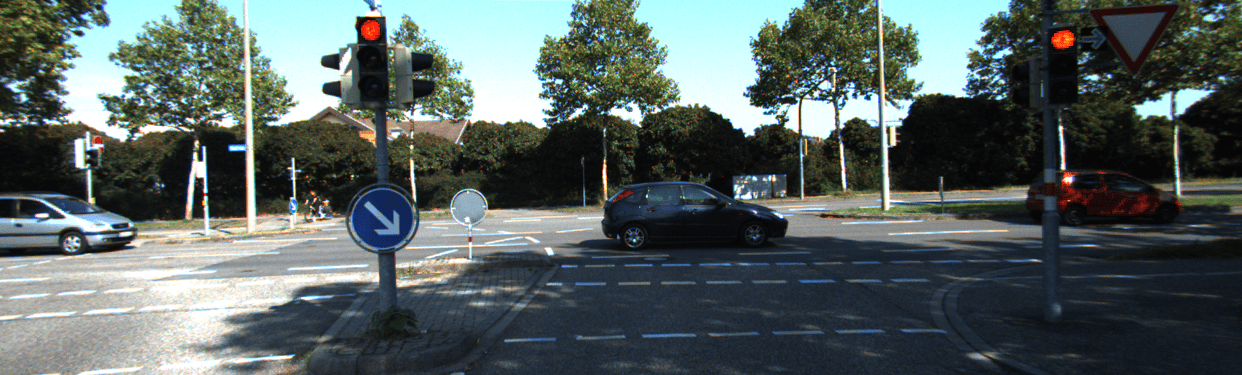} } 
\subfigure{
\includegraphics[width=0.32\linewidth]{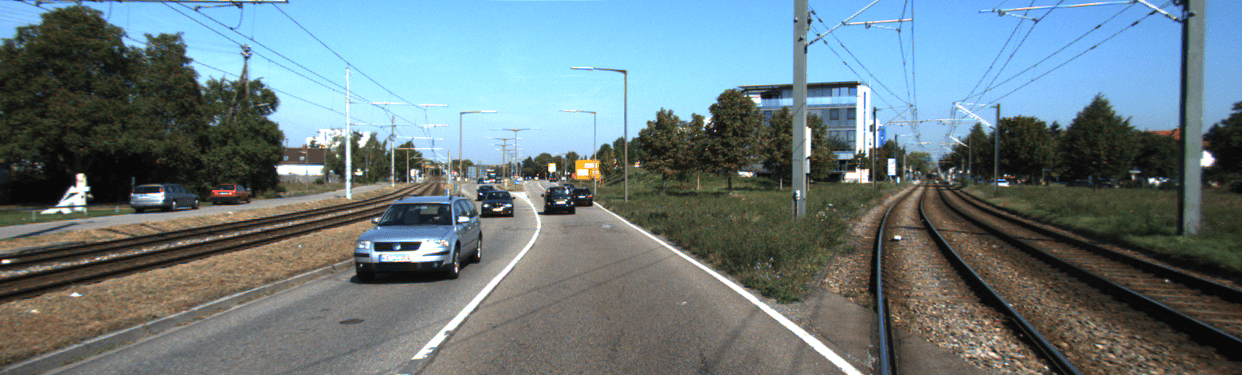} } 
\subfigure{
\includegraphics[width=0.32\linewidth]{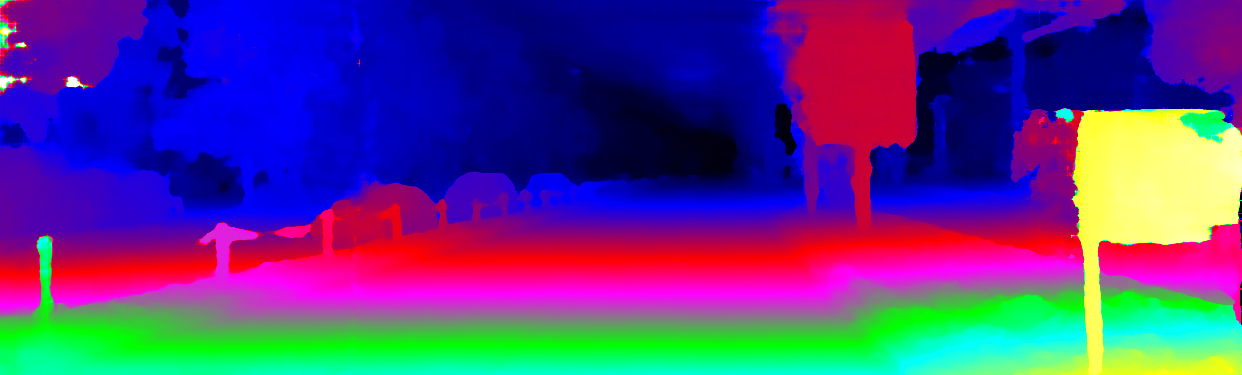} } 
\subfigure{
\includegraphics[width=0.32\linewidth]{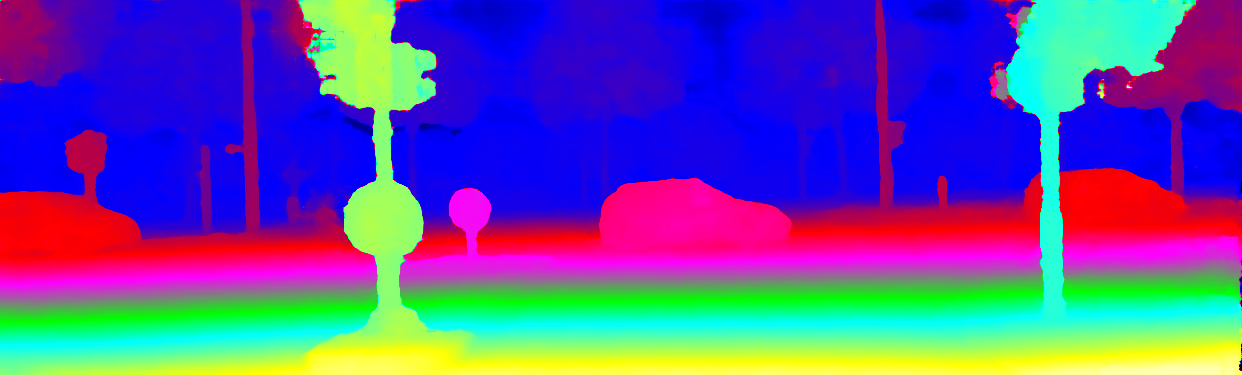} } 
\subfigure{
\includegraphics[width=0.32\linewidth]{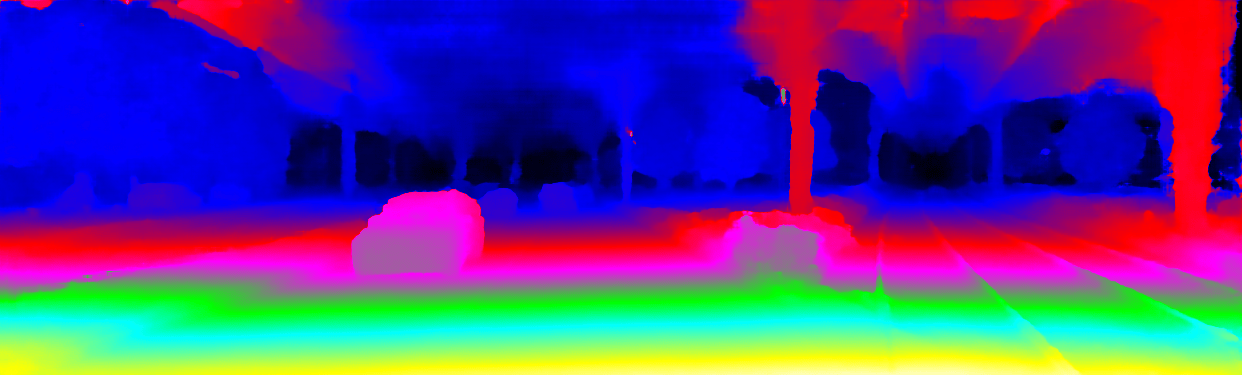} } 
\subfigure{
\includegraphics[width=0.32\linewidth]{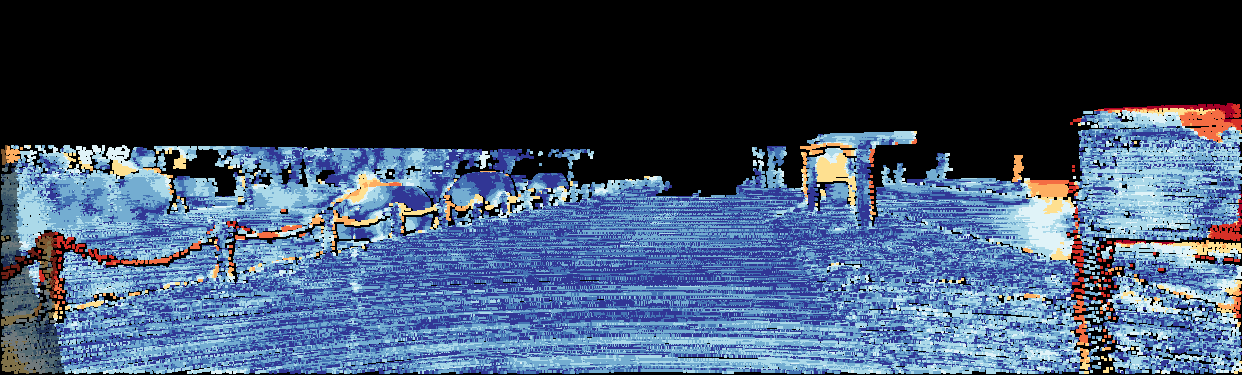} } 
\subfigure{
\includegraphics[width=0.32\linewidth]{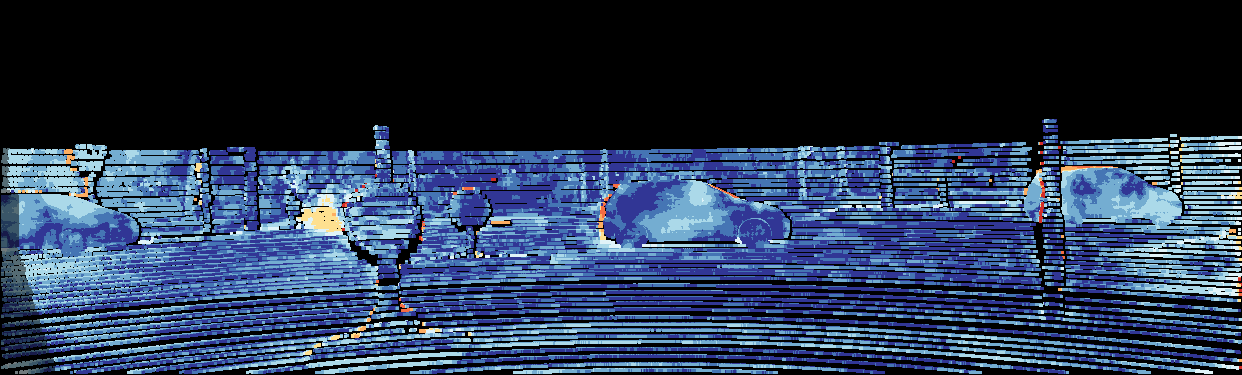} } 
\subfigure{
\includegraphics[width=0.32\linewidth]{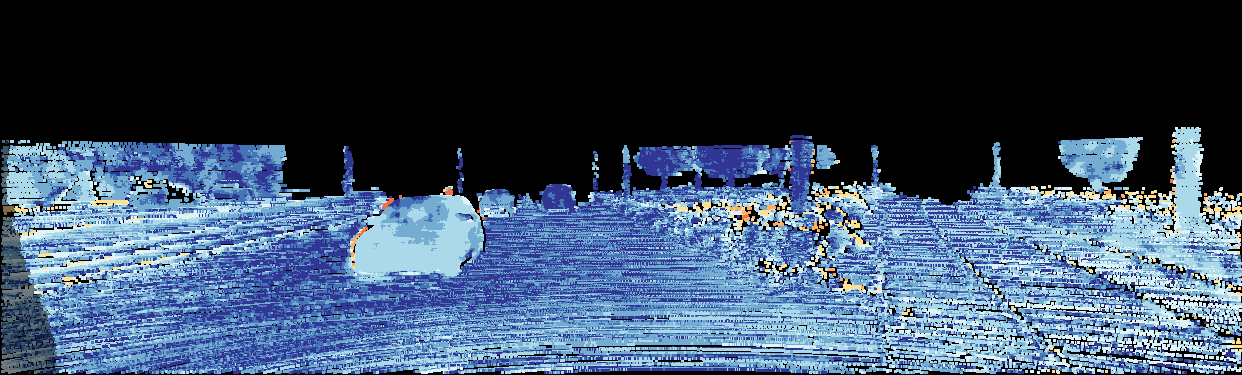} } 
\subfigure{
\includegraphics[width=0.32\linewidth]{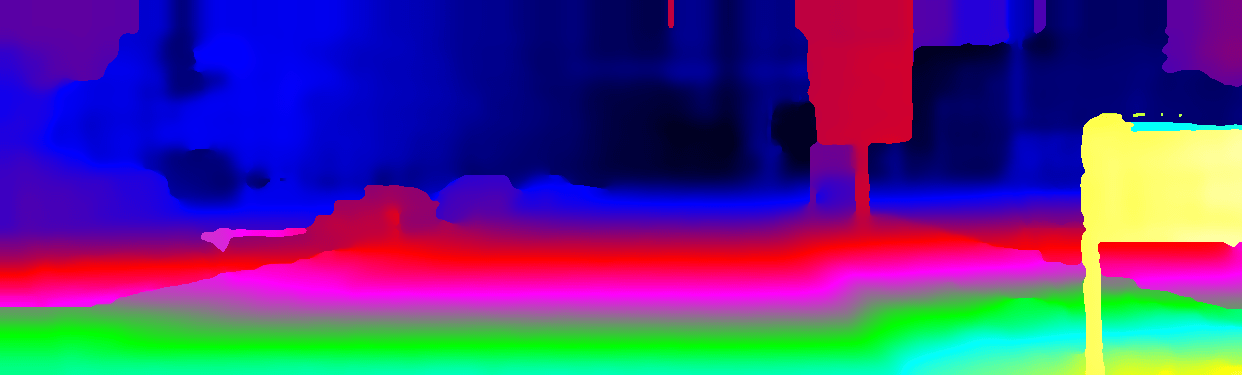} } 
\subfigure{
\includegraphics[width=0.32\linewidth]{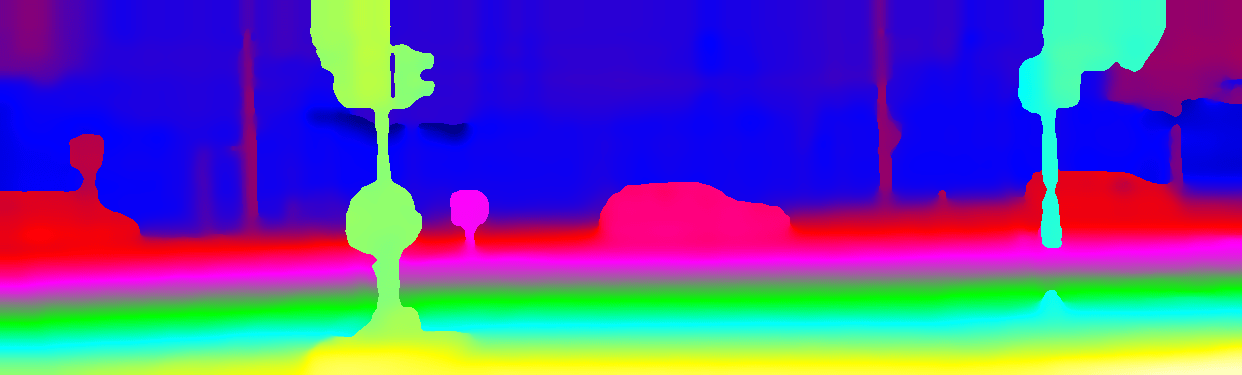} } 
\subfigure{
\includegraphics[width=0.32\linewidth]{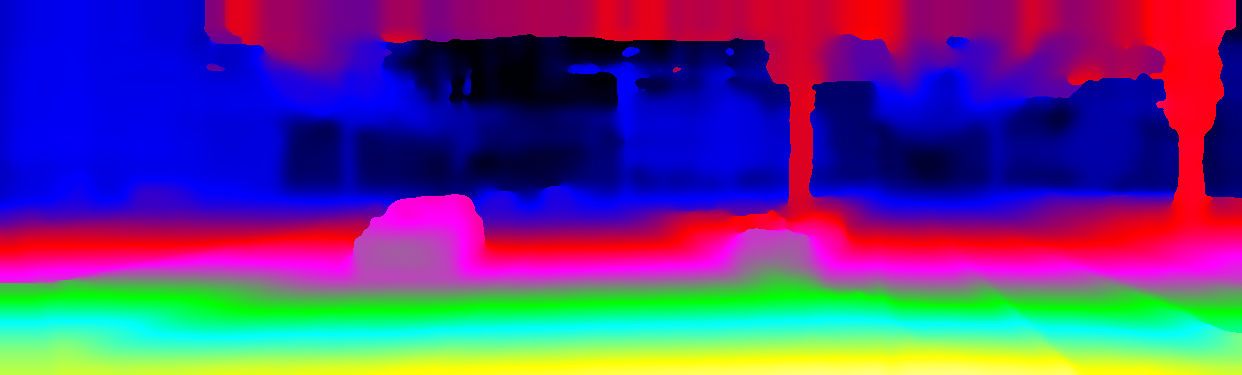} } 
\subfigure{
\includegraphics[width=0.32\linewidth]{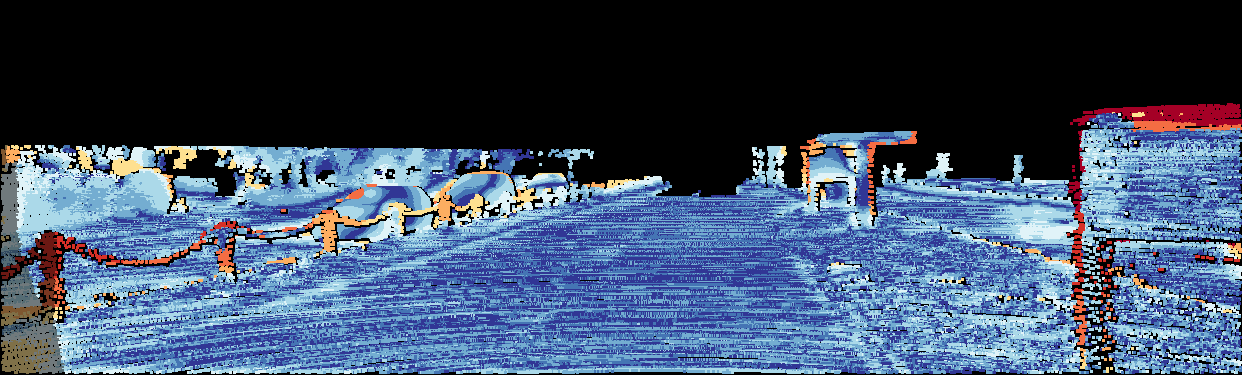} } 
\subfigure{
\includegraphics[width=0.32\linewidth]{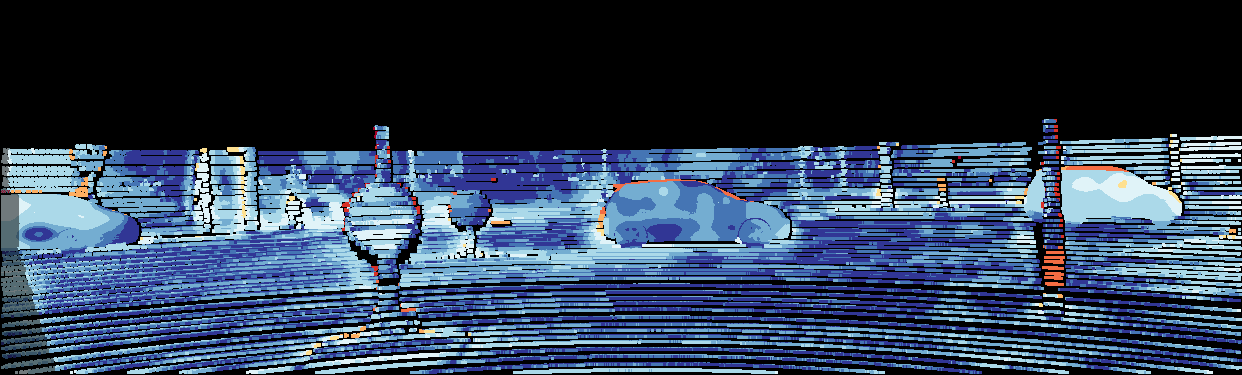} } 
\subfigure{
\includegraphics[width=0.32\linewidth]{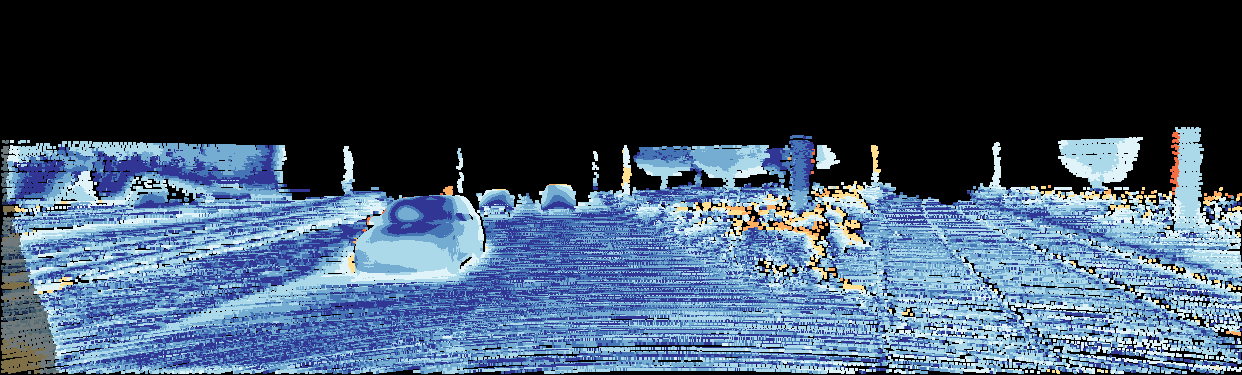} } 
\caption{\label{fig:kitti15}\textbf{Our qualitative results on KITTI-2015:} Top to bottom: left image, our result, our error map, result of MC-CNN-acrt \cite{Zbontar2016} and its error map.}
\end{center}
\end{figure*}

There is a subtle but important difference between KITTI 2012 and 2015: in KITTI 2015, CAD models are inserted in place of moving cars so that vehicles are densely labeled. As a consequence, highly reflected areas such as car glass are included in the evaluation. This leads to a bias in evaluating the stereo matching performance as vehicles consume the majority of weights in evaluation and the actual depth value of the window instead of the real disparity value is selected for ambiguous disparity values on transparent surfaces. In Fig.~\ref{fig:kitti12} and Fig.~\ref{fig:kitti15} we show qualitative results of our method and comparison with MC-CNN \cite{Zbontar2016} on KITTI 2012 and KITTI 2015 datasets.




\begin{table}[!htp]
\tabcolsep=0.09cm
\begin{tabularx}{\linewidth}{c | c | c | c | c }
{\bf Method} & {\bf Out-Noc} & {\bf Out-All} & {\bf Avg-Noc} & {\bf Avg-All} \\ \hline
GC-NET\cite{KendallMDHKBB17}  & 2.71 \% & 3.46 \% & 0.6 px & 0.7 px  \\
Displets v2\cite{GuneyG15}  & 3.43 \% & 4.46 \% & 0.7 px & 0.8 px \\
SGM-Net\cite{Seki2017CVPR}  & 3.60 \% & 5.15 \% & 0.7 px & 0.9 px \\
PBCP\cite{Seki2016BMVC} & 3.62 \% & 5.01 \% & 0.7 px & 0.9 px \\
L-ResMatch\cite{shaked2016stereo}  & 3.64 \% & 5.06 \% & 0.7 px & 1.0 px \\
MC-CNN-acrt\cite{Zbontar2016}  & 3.90 \% & 5.45 \% & 0.7 px & 0.9 px \\
SPS-St\cite{Yamaguchi14} & 4.98 \% & 6.28 \% & 0.9 px & 1.0 px \\ \hline
SsSMnet & 3.34 \% & 4.24 \% & 0.7 px & 0.8 px \\ \hline
\end{tabularx}
\caption{\label{tab:kitti12}\textbf{Results on KITTI 2012 stereo benchmark (as of 3 September 2017).}}
\end{table}

\begin{table}[!htp]
\tabcolsep=0.15cm
\begin{tabularx}{\linewidth}{c | c | c | c | c }
{\bf Method} & {\bf D1-bg} & {\bf D1-fg} & {\bf D1-all} & {\bf Runtime} \\ \hline
CRL\cite{pang2017cascade} & 2.48 \% & 3.59 \% & 2.67 \% & 0.47 s \\
GC-NET\cite{KendallMDHKBB17} & 2.21 \% & 6.16 \% & 2.87 \%  & 0.9 s \\
SGM-Net\cite{Seki2017CVPR} & 2.66 \% & 8.64 \% & 3.66 \%  & 67 s \\
L-ResMatch\cite{shaked2016stereo} & 2.72 \% & 6.95 \% & 3.42 \% & 48 s\\
MC-CNN-acrt\cite{Zbontar2016}  & 2.89 \% & 8.88 \% & 3.89 \% & 67 s\\ 
Displets v2\cite{GuneyG15} & 3.00 \% & 5.56 \% & 3.43 \% & 265 s \\ \hline 
SsSMnet & 2.86 \% & 7.12 \% & 3.57 \%  & 0.8 s\\ \hline
\end{tabularx}
\caption{\label{tab:kitti15}\textbf{Results on KITTI 2015 stereo benchmark ((as of 3 September 2017)}}
\end{table}

\subsection{Middlebury}
The stereo pairs in the Middlebury stereo dataset are indoor scenes with multiple handcrafted layout. The ground truth disparities are captured by structured light with higher density and precision than KITTI dataset. We select 18 pairs out of 31 from Middlebury 2001 \cite{Scharstein2002} 2002 \cite{Scharstein2003} 2005 \cite{Scharstein07} and 2006 \cite{HirschmullerS07} to evaluate the generalization ability among current state-of-the-art learning free conventional method SPS-st \cite{Yamaguchi14} and deep learning based method MC-CNN \cite{Zbontar2016}. We also compare our method with the state-of-the-art traditional method on Middlebury benchmark, MeshStereo \cite{Zhangiccv15}, as a reference to highlight our performance. 

\begin{table*}[!htp]
\centering
\scriptsize
\tabcolsep=0.1225cm
\begin{tabularx}{\textwidth}{c|c|cccccccccccccccccc|c}
Method                       & \rot{Threshold} & \rot{Venus} & \rot{Dolls} & \rot{Laundry} & \rot{Moebius} & \rot{Reindeer} &\rot{Aloe} &\rot{Baby1} &\rot{Baby2} &\rot{Baby3} &\rot{Cloth1} &\rot{Cloth2} &\rot{Cloth3} &\rot{Cloth4} &\rot{Rocks1} &\rot{Rocks2} &\rot{Tsukuba} & \rot{ConesH} & \rot{TeddyH}  & Mean \\  \hline

\multirow{2}{*}{SPS-St}      & 0.5       & 9.31 & 37.17 & 30.02 & 31.98 & 21.67 & 18.24 & 9.11 & 14.55 & 15.63 & 4.85 & 18.56 & 18.95 & 11.95 & 19.96 & 14.49 & 33.69 & 23.63 & 26.97 & 20.04\\ \cline{2-21} 
                             & 1         & 4.38 & 15.54 & 18.69 & 17.38 & 11.05 & 8.57 & 3.01 & 5.06 & 6.38 & 0.63 & 6.17 & 6.15 & 4.00 & 6.57 & 5.23 & 12.83 & 5.91 & 10.86 & 8.25   \\ \hline
\multirow{2}{*}{MC-CNN-arct} & 0.5       & 16.57 & 53.60 & 37.92 & 42.11 & 34.17 & 31.11 & 17.75 & 25.72 & 27.34 & 27.25 & 49.68 & 43.03 & 42.10 & 36.12 & 42.34 & 31.08 & 30.63 & 31.85  & 34.47 \\ \cline{2-21} 
                             & 1         & 5.70 & 23.78 & 25.36 & 20.96 & 14.57 & 16.72 & 7.93 & 13.43 & 10.62 & 7.70 & 24.93 & 15.02 & 10.94 & 12.84 & 13.77 & 17.40 & 10.23 & 15.13 & 14.84 \\ \hline
\multirow{2}{*}{SsSMnet}     & 0.5       & 7.27 & 17.68 & 25.79 & 18.48 & 16.46 & 10.87 & 9.07 & 13.49 & 11.29 & 1.09 & 6.90 & 5.10 & 5.30 & 12.69 & 5.82 & 37.16 & 16.85 & 24.76 & 13.67\\ \cline{2-21} 
                             & 1         & 2.86 & 7.58 & 15.93 & 12.27 & 9.30 & 5.67 & 4.32 & 4.00 & 6.16 & 0.42 & 2.64 & 2.52 & 3.08 & 2.94 & 2.69 & 11.90 & 5.10 & 9.90 & 6.07\\ \hline
                              \hline
\multirow{2}{*}{MeshStereo}      & 0.5       & 7.88 & 29.64 & 29.67 & 23.75 & 16.77 & 17.09 & 11.45 & 14.11 & 16.42 & 5.88 & 18.47 & 14.20 & 13.83 & 18.10 & 13.16 & 31.49 & 21.01 & 21.26 & 18.01\\ \cline{2-21} 
                             & 1         & 1.04 & 11.59 & 16.61 & 14.19 & 6.98 & 9.57 & 3.73 & 3.13 & 6.39 & 1.84 & 6.99 & 4.21 & 4.97 & 5.68 & 3.49  & 12.80 & 3.71 & 8.30 &  6.96  \\ \hline
\end{tabularx}
\caption{\label{tab:mb}\textbf{Cross datasets performance on Middlebury stereo dataset.} Baseline methods are using the same parameters released by the authors. We test MC-CNN model trained on KITTI for a fair comparison. Our method updates parameters in an on-line way and we show the results after 100 iterations. Note: MeshStereo is tuned on the Middlebury dataset.}
\end{table*}

\begin{figure*}[!htp]
\begin{center} 
\subfigure{
\includegraphics[width=\linewidth]{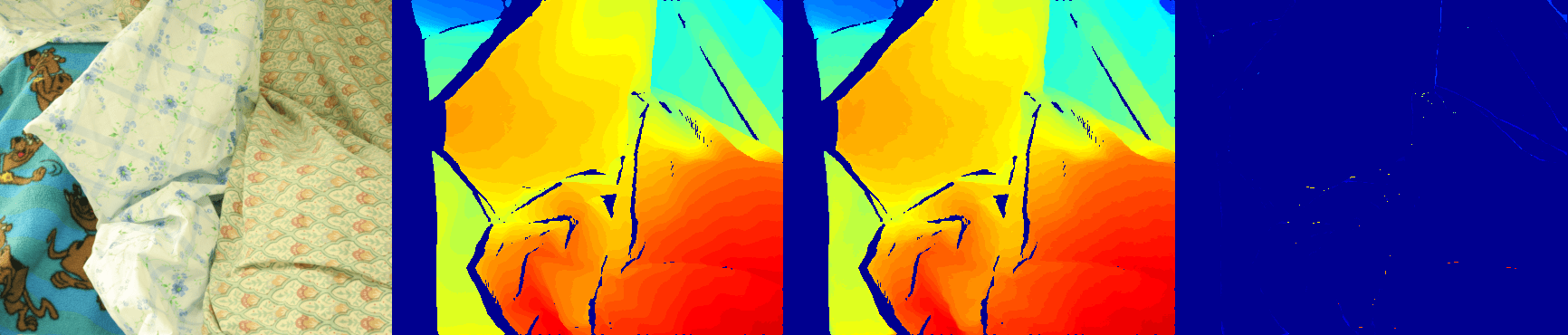} }\vspace{-0.2cm} 
\subfigure{
\includegraphics[width=\linewidth]{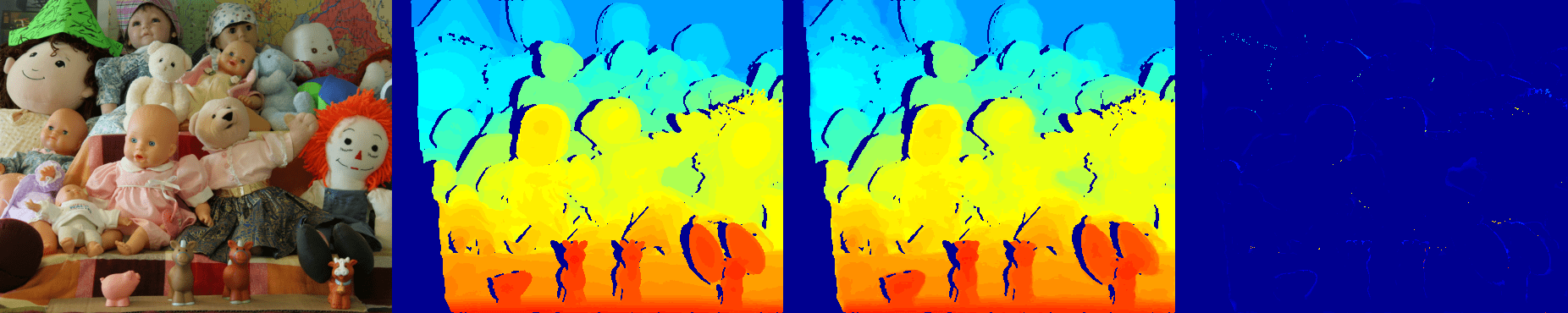} }\vspace{-0.2cm}
\subfigure{
\includegraphics[width=\linewidth]{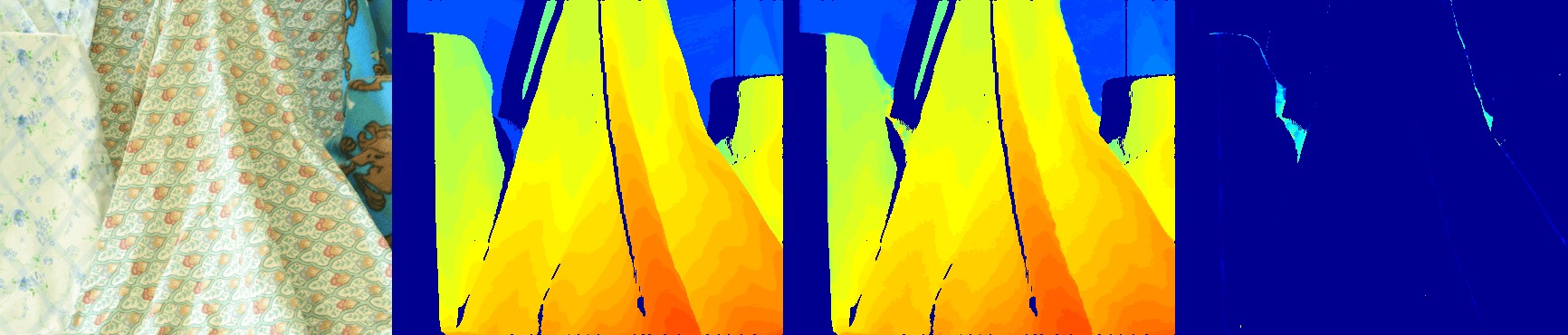} }\vspace{-0.2cm} 
\subfigure{
\includegraphics[width=\linewidth]{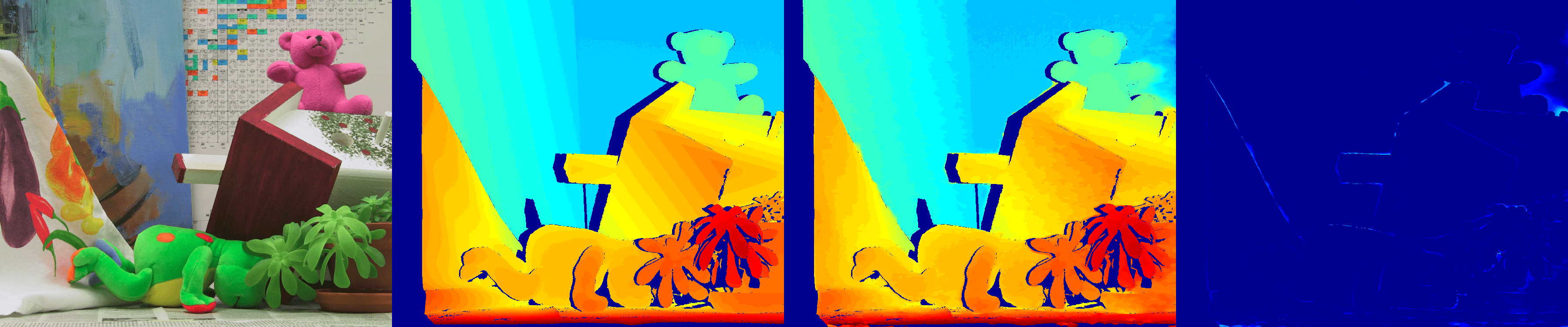} }\vspace{-0.2cm} 
\subfigure{
\includegraphics[width=\linewidth]{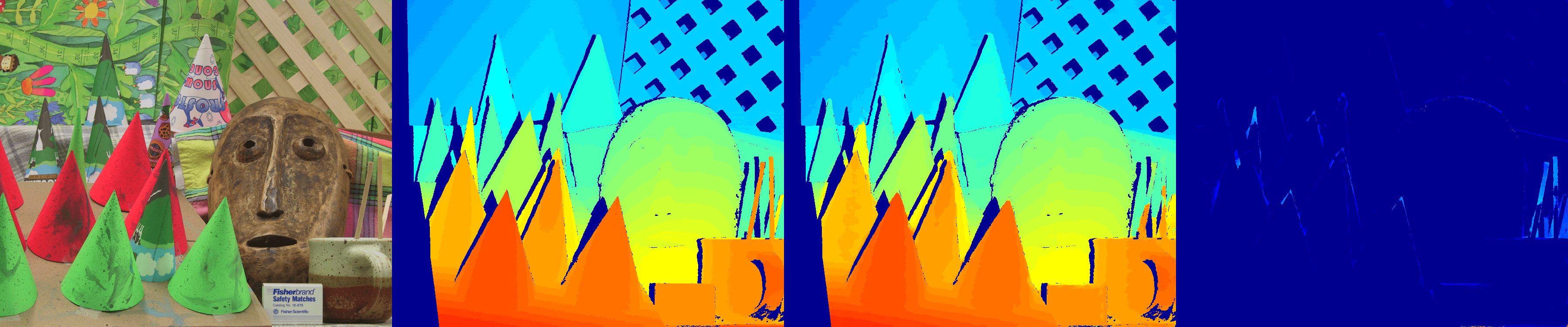} }
\caption{\label{fig:result_mb_e}\textbf{Our qualitative results on Middlebury:} Left to right: left image, ground truth disparity map, our result, error map.}
\end{center}
\end{figure*}

For SPS-st \cite{Yamaguchi14} and MeshStereo \cite{Zhangiccv15}, we use the same parameters and code released by the authors. For MC-CNN, we use the model trained on the KITTI dataset and the post-processing parameters tuned on the KITTI dataset as well.  As shown in Table~\ref{tab:mb}, our method outperforms all baseline methods with a notable margin. Our method achieves an improvement of $46.60\%$ and $152.16\%$ with threshold 0.5 pixel on none-occluded pixels compared with SPS-st and MC-CNN respectively. For the conventional method that tuned parameters on the Middlebury dataset, our method performs $31.75\%$ and $14.66\%$ better with $0.5$ and $1$ pixel threshold respectively. All experiments are evaluated on third-size of original resolution due to the limited size of the GPU's memory available except for Middlebury 2002. We run it on the half-size resolution. Some of our qualitative results are shown in Fig.~\ref{fig:result_mb_e}.

According to these results, we would like to advocate that although traditional learning free methods claim they are suitable for general cases, they still need to manually tune the meta parameters for different datasets. Supervised deep-learning based methods seriously suffer from dataset sensitivity. Our method, on the other hand, is able to self-adapt to different scenarios. 

\section{Conclusion}
We have presented a new deep stereo matching network that can be trained end-to-end using the input stereo image pairs only, without the need of ground-truth depth maps.  A novel training loss is proposed to exploit the loop constraint in image warping process and to handle the textureless areas. Our network can be run in an on-line learning fashion when being exposed to new, never-seen-before images, and it can self-improve by adapting itself to the new imageries, as no ground-truth labeling is needed. Experiments show the method achieves superior performance than traditional learning-free methods as well as recent supervised deep-learning based methods. In future, we plan to explore occlusion reasoning in order to better handle visual occlusion.  

\noindent
\textbf{Acknowledgement.\ } We gratefully acknowledge the support of NVIDIA Corporation with donation of TITAN Xp GPU used for this research, as well a NVIDIA Drive-PX2 platform for an autonomous driving project.  YZ's PhD scholarship is funded by CSIRO Data61. YD is supported in part by ARC DECRA project (DE140100180). This work is funded in part by ARC Centre of Excellence for Robotic Vision (ARC-ACRV-CE14). 

{\small
\bibliographystyle{ieee}
\bibliography{Stereo-Matching-Reference}
}

\end{document}